%% file: main.tex
\documentclass[lettersize,journal]{IEEEtran}
\usepackage{amsmath,amsfonts}
\usepackage{algorithmic}
\usepackage{multirow}
\usepackage{rotating}
\usepackage{algorithm}
\usepackage{array}
\usepackage[caption=false,font=normalsize,labelfont=sf,textfont=sf]{subfig}
\usepackage{textcomp}
\usepackage{stfloats}
\usepackage{hyperref}
\usepackage{verbatim}
\usepackage{graphicx}
\usepackage{cite}
\usepackage{color}
\usepackage{booktabs}
\usepackage[capitalise]{cleveref}
\crefname{section}{Sec.}{Sec.}
\usepackage{amssymb}
\begin{document}

\title{Motion-aware Memory Network for \\ Fast Video Salient Object Detection}

\author{\IEEEauthorblockN{
Xing~Zhao,
Haoran~Liang,
Peipei~Li,
Guodao~Sun,
Dongdong~Zhao,
Ronghua~Liang,~\IEEEmembership{Senior~Member,~IEEE},\\
and Xiaofei~He,~\IEEEmembership{Senior~Member,~IEEE}
}
\thanks{Xing Zhao, Haoran Liang, Dongdong Zhao, Guodao Sun, Peipei Li, and Ronghua Liang are with Zhejiang University of Technology, Hangzhou 310023, China (e-mail: \{xing, haoran, zhaodd, guodao, peipeili, rhliang\}@zjut.edu.cn).\textit{(Corresponding author: Haoran~Liang.)}}
\thanks{Xiaofei He is with Zhejiang University, Hangzhou 310023, China (e-mail: xiaofeihe@cad.zju.edu.cn).}
\thanks{Code available at \url{https://github.com/zhaoxing2022/MMN-VSOD}.}
}



\maketitle

\begin{abstract}
Previous methods based on 3DCNN, convLSTM, or optical flow have achieved great success in video salient object detection (VSOD).
However, these methods still suffer from high computational costs or poor quality of the generated saliency maps.
To address this, we design a space-time memory (STM)-based network that employs a standard encoder--decoder architecture.
During the encoding stage, we extract high-level temporal features from the current frame and its adjacent frames, which is more efficient and practical than methods reliant on optical flow.
During the decoding stage, we introduce an effective fusion strategy for both spatial and temporal branches.
The semantic information of the high-level features is used to improve the object details in the low-level features.
Subsequently, spatiotemporal features are methodically derived step by step to reconstruct the saliency maps.
Moreover, inspired by the boundary supervision prevalent in image salient object detection (ISOD), we design a motion-aware loss that predicts object boundary motion, and simultaneously perform multitask learning for VSOD and object motion prediction.
This can further enhance the model's capability to accurately extract spatiotemporal features while maintaining object integrity.
Extensive experiments on several datasets demonstrate the effectiveness of our method and can achieve state-of-the-art metrics on some datasets.
Our proposed model does not require optical flow or additional preprocessing, and can reach an impressive inference speed of nearly 100 FPS\@.

\end{abstract}


\input{1introduction}
\input{2relatedwork}
\input{3method}

\input{4experiments}

\input{5conclusion}

\section*{Acknowledgments}
This work is partially supported by the National Key Research and Development Program of China (2020YFB1707700), the National Natural Science Foundation of China (62176235, 62036009), and Zhejiang Provincial Natural Science Foundation of China (LY21F020026).



\bibliographystyle{IEEEtran}
\bibliography{main}

\vfill

\end{document}

%% file: 1introduction.tex
\section{Introduction}\label{sec:introduction}
\IEEEPARstart{s}{hort} videos have gained immense popularity in recent years.
However, analyzing and processing vast amounts of video data remains a formidable challenge.
As a fundamental task in computer vision, VSOD aims to identify and segment the most visually striking objects within videos;
it has many crucial applications and benefits the downstream tasks in practical scenarios, such as video compression~\cite{hadizadeh2013saliency}, video editing~\cite{jain2016click}, video object tracking~\cite{wu2014weighted}, and person re-identification~\cite{zhao2016person}.
Compared with ISOD task, VSOD places a greater emphasis on the fusion of temporal information.
Given the limited diversity of categories in video training datasets, VSOD proves to be more challenging than ISOD.
\begin{figure}[t]
	\centering
	\includegraphics[width=0.5\textwidth]{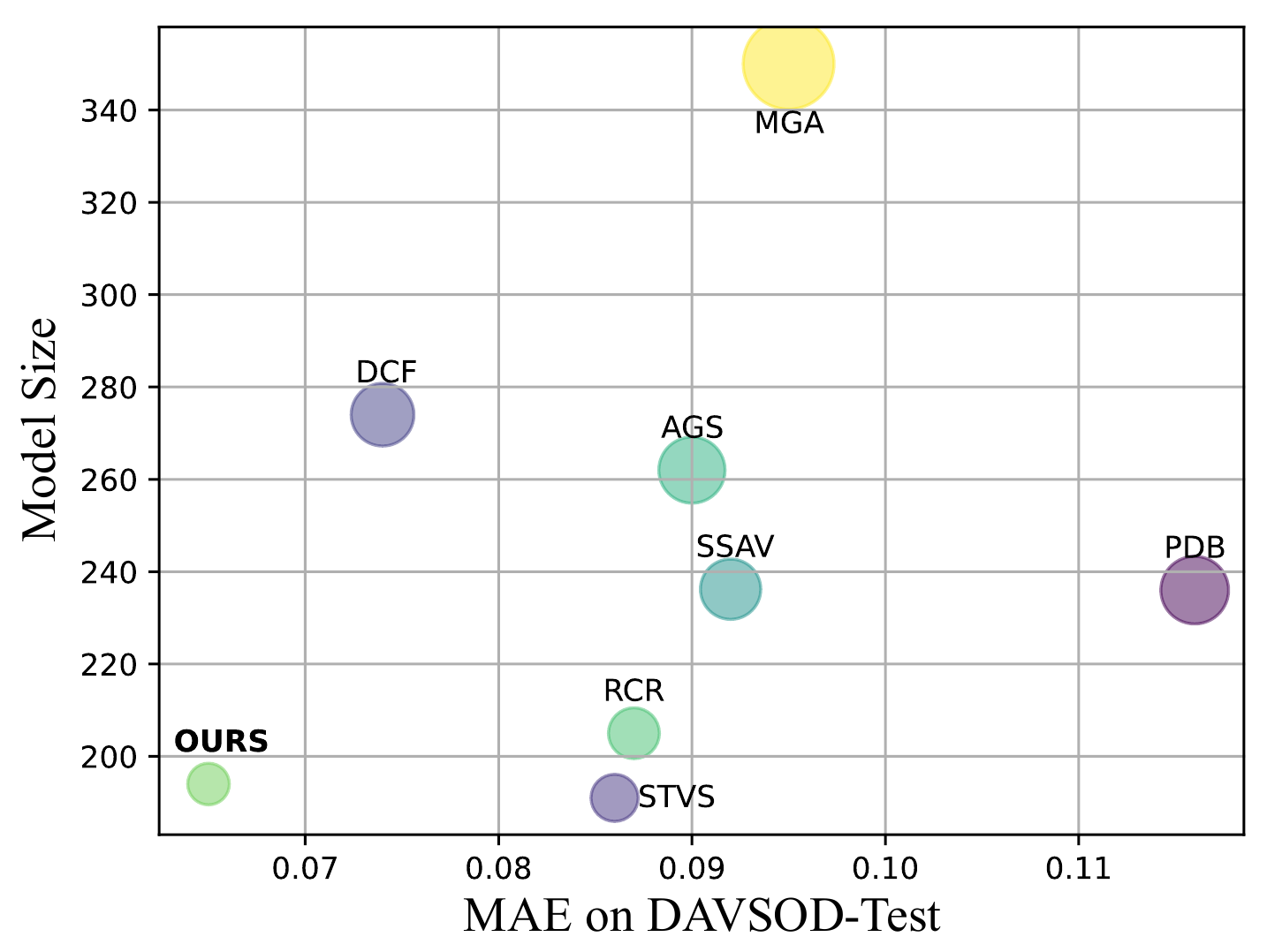}
	\caption{Comparison of model size and mean absolute error (MAE) on DAVSOD. Models situated closer to the bottom-left corner are more efficient and effective. DCF~\cite{zhang2021dynamic} and STVS~\cite{chen2021exploring} represent the latest methodologies, the former has a better metric in terms of MAE but contains numerous parameters, while the latter has fewer parameters but compromises on its MAE performance.
	Our method achieves the best MAE metric on the largest VSOD dataset while utilizing fewer parameters. Moreover, its inference speed surpasses that of the majority of prior methods.}
	\label{fig:fig1}
\end{figure}

3DCNN and convLSTM are predominantly employed to discern precise temporal salient cues in the early stages.
3DCNN, burdened with a large number of parameters, is challenging to optimize~\cite{jiao2021guidance}.
Furthermore, its proficiency in extracting temporal details falls short when compared to optical flow.
Additionally, its spatial information extraction capability doesn't match that of 2DCNN pre-trained on large-scale datasets.
ConvLSTM struggles to harness spatial multilevel features for an accurate reconstruction of the final saliency prediction~\cite{chen2021exploring}, and the results it generates tend to be blurry.
Recently, methods that capitalize on optical flow have demonstrated notable advancements in VSOD, but they come with the drawback of significantly increased computational demands.
FlowNet-2.0~\cite{ilg2017flownet}, a well-known model used for generating optical flow, has a model size and computational cost that exceeds those of many VSOD methods.
The preprocessing to obtain optical flow may pose challenges when applied in practice.

The STM-based network has demonstrated its robust capabilities in capturing temporal information for video object segmentation (VOS) tasks.
Different from VOS, which segments a consistent object, VSOD not only needs to segment salient objects, but also identify the most attractive objects within the current frame.
Directly applying STM for VSOD might face challenges in scenarios where salient objects are constantly changing, namely, saliency shift issue~\cite{fan2019shifting} in VSOD\@.
Recognizing the significance of high-level features in distinguishing salient objects, we propose an adjacent space-time memory module (ASTM) built upon high-level features.
This guides our model to find salient objects and accurately gather temporal information from adjacent frames without optical flow acquisition.
In our model, the ASTM serves as the temporal branch of the model.
It solely processes the temporal information associated with the current frame from its adjacent counterparts.
Conversely, the spatial branch handles the spatial information of the present frame.
The two branches subsequently merge in the decoder.

In the decoding stage, low-level features play a crucial role in reconstructing the final prediction.
As indicated in~\cite{fan2019shifting}, the absence of these low-level features leads to extremely small saliency maps, thereby causing blurriness upon upsampling.
We design an effective fusion method for multilevel features from the spatial and channel perspectives.
Moreover, drawing inspiration from the boundary supervision employed in ISOD, we have integrated a comparable boundary motion supervision for VSOD.
This approach is then combined with VSOD in a multitask learning framework, facilitating mutual enhancement.
As illustrated in \cref{fig:fig1}, our model outperforms others on the largest VSOD dataset, DAVSOD, in terms of MAE, and the model size is smaller than most previous methods.

\begin{figure*}[ht!]
	\centering
	\includegraphics[width=1\textwidth]{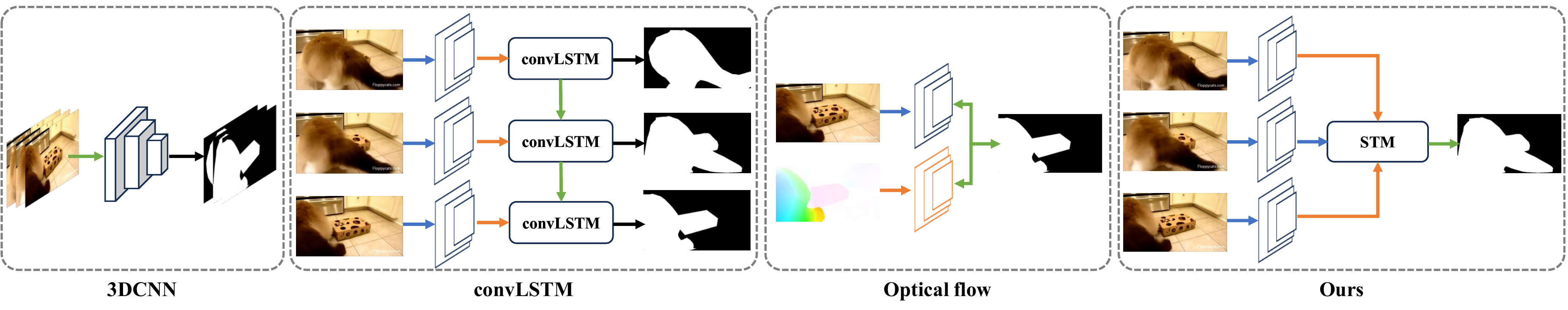}
	\caption{Comparative overview of the architecture of different methods.}
	\label{fig:diff-methods}
\end{figure*}

The main contributions of this article can be summarized as follows:
\begin{enumerate}
\item While conventional VSOD typically comprises two branches, we incorporate the STM mechanism into VSOD, utilizing it as a temporal branch.
This strategy offers an efficient alternative to previous methods that were either computationally burdensome or complex to implement.
Our proposed ASTM effectively gathers salient object information across adjacent frames, subsequently providing decisive guidance to the decoder in generating accurate predictions.

\item We propose an efficient fusion strategy to encourage low- and high-level features cooperated between low- and high-level features, regardless of whether they belong to the temporal or spatial branch.
The semantic information from the high-level features guides the detailed object information within the low-level features, refining the predicted maps step by step.
\item We extend boundary supervision commonly used by ISOD as multitask learning into VSOD, and a motion-aware loss is designed to enable the model to focus on object motion while maintaining object integrity.
\item The experimental results validate the strong performance of our proposed method across various well-known VSOD datasets, with particularly notable results on the highly challenging DAVSOD dataset.
Furthermore, our model's inference speed is over twice as fast as the state-of-the-art method, DCF~\cite{zhang2021dynamic}.
\end{enumerate}

%% file: 2relatedwork.tex
\begin{figure*}[ht]
    \centering
    \includegraphics[width=1\textwidth]{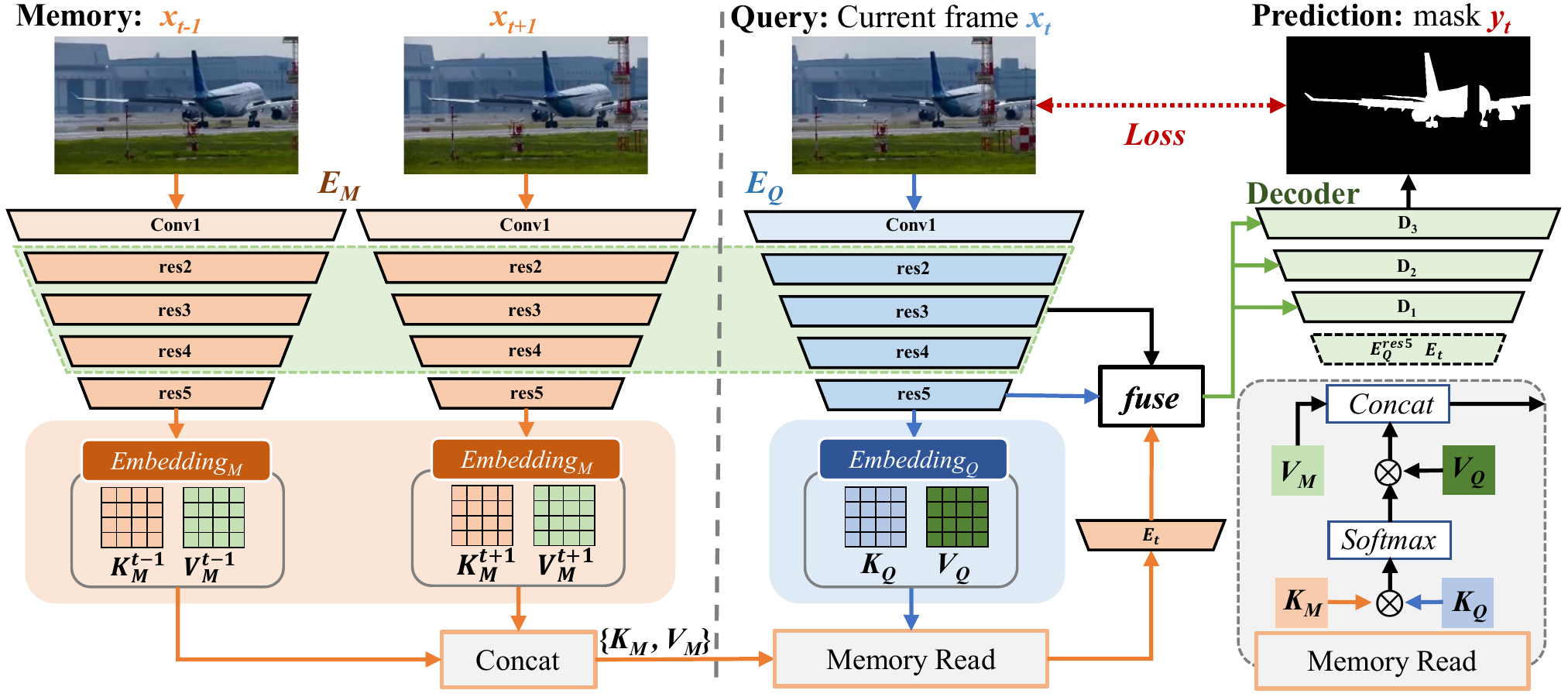}
    \caption{Overall structure of the model. Two adjacent frames [$x_{t-1},x_{t+1}$] are placed into the memory when the first frame $x_t$ is sent to the encoder. High-level temporal features ($E_t$) are obtained by the memory read operation. Finally, two high-level features $E_t$ (temporal branch) and $E_Q^{res5}$ (spatial branch) and all the low-level features are used to reconstruct the final saliency map.}
    \label{fig:model}
\end{figure*}

\section{related work}\label{sec:related-work}
In this section, we initially provide an overview of the development of VSOD from early stage to the present, and then introduce the memory network and its variants that are commonly used in VOS\@.

\subsection{Video salient object detection}\label{subsec:video-salient-object-detection}

Prior to the boosting of deep learning, several methods have achieved encouraging progress by using hand-engineered features. Wang \textit{et al.}~\cite{wang2015consistent} incorporated the boundary information of a single frame and the motion information between frames to predict salient regions based on the gradient flow field.
Xi \textit{et al.}~\cite{xi2016salient} extended background priors to videos and integrated a bidirectional consistency propagation to obtain spatiotemporal background priors.
Liu \textit{et al.}~\cite{liu2014superpixel} measured temporal saliency by extracting motion and color histograms at the superpixel level.
Xu \textit{et al.}~\cite{xu2016learning} conducted statistical analysis on their eye-tracking database and integrated the most relevant HEVC features for video saliency detection.
These methods are less effective to deep learning-based methods given their inability to extract high-level features.

VSOD has also gradually shifted to deep learning-based approaches due to the tremendous progress of deep learning in recent years.
Wang \textit{et al.}~\cite{wang2017video} trained a static image saliency model with a fully convolutional network prior to guiding subsequent video predictions.
Li \textit{et al.}~\cite{li2018flow} combined optical flow and convLSTM to obtain temporal coherence at the feature level; the flow was dynamically updated rather than pre-computed.
Song \textit{et al.}~\cite{song2018pyramid} designed a dilated module to extract spatial features at multiscale; these features are then sent into a bidirectional convLSTM network to learn spatiotemporal information.
Different from previous works that learned temporal information between adjacent frames, Chen \textit{et al.}~\cite{chen2019improved} utilized long-term information to improve VSOD by aligning non-local frames.
Le \textit{et al.}~\cite{le2018video} employed 2D and 3DCNN to learn local and global features simultaneously, and the computation costs were significantly higher.
Li \textit{et al.}~\cite{li2019motion} also introduced a two-branch network, one for still image detection and the other for motion detection with optical flow.
Compared with 3DCNN or convLSTM, this two-branch architecture has become increasingly popular lately.

VSOD has always been a difficult task due to the lack of large-scale datasets.
Fan \textit{et al.}~\cite{fan2019shifting} released the largest dataset DAVSOD that contained more realistic scenes and higher quality object masks.
Zhao \textit{et al.}~\cite{zhao2021weakly} also published a scribble version of DAVSOD for semi-supervised learning.
The emergence of DAVSOD has alleviated this problem to a certain degree, and more effective methods are proposed subsequently.

Gu \textit{et al.}~\cite{gu2020pyramid} applied a self-attention mechanism to capture long-range information to reduce computation and memory cost.
They designed pyramid structures to find object trajectory regions and then measured the pixel-wise relation.
Instead of using 3DCNN at the encoder, Chen \textit{et al.}~\cite{chen2021exploring} utilized it to capture motion sensing at the decoder.
Zhang \textit{et al.}~\cite{zhang2021dynamic} used dynamic convolution kernels to sense dynamic changing scenes.

Off-the-shelf optical flow models also yield rich temporal information.
This two-branch architecture has achieved a new level of accuracy when cooperating with static image branch.
Ren \textit{et al.}~\cite{ren2020tenet} designed a curriculum learning strategy from spatial and temporal excitations.
They proposed an online excitation process to refine the saliency maps recurrently in the testing stage.
Ji \textit{et al.}~\cite{ji2021full} considered the bidirectional flow of temporal and spatial information when fusing cross-modal features.

Our model also follows a two-branch structure, but we directly learn temporal information from frames rather than optical flows that may bring high computation costs.
\cref{fig:diff-methods} shows a comparative overview of the architecture of several methods.

\subsection{Memory network on VOS}\label{subsec:2.3}

The VOS methods mentioned in this article are semi-supervised.
That is, only the mask of the first frame is provided, and all masks should be predicted according to the subsequent frames.
High-level features are always discarded on VOS because the target object will not change during the whole process, the model does not need high-level semantic information to distinguish different objects, and a large feature size can be retained.

The memory network was first used in the field of natural language processing~\cite{kumar2016ask,miller2016key,sukhbaatar2015end}, and later applied to some computer vision tasks~\cite{na2017read,chunseong2017attend,yang2018learning}.
Oh \textit{et al.}~\cite{oh2019video} initially introduced STM to the VOS task, the previous frames were encoded to external memory, and the current frame would be segmented by querying the relevant information from the previous memory.
They achieved state-of-the-art results at that time and promoted the development of other variants of STM in VOS\@.

STM was non-local, but the target object occupied only a local part of the image which would lead to a mismatch problem.
Seong \textit{et al.}~\cite{seong2020kernelized} presented a kernelized memory read operation to conduct Query-to-Memory and Memory-to-Query matching to solve the mismatch problem.
They also explored multiple kernel types and suitable kernel selection with learnable parameters in~\cite{seong2022video}.
Xie \textit{et al.}~\cite{xie2021efficient} memorized the local regions, where the target object appeared in previous frames, and then conducted regional memory read efficiently and effectively by local-to-local matching.
In addition to pixel-level matching, Hu \textit{et al.}~\cite{hu2021learning} combined pixel-level and object-level information to learn position consistency among frames.

VOS is relatively similar to VSOD, but the target object in VSOD may be one or more; it is even constantly changing in the temporal domain.
We apply STM to VSOD for the first time and improve it to become more suitable for our task.

%% file: 3method.tex
\section{Method}\label{sec:method}
In this section, we begin by outlining the overall architecture of the network, followed by a discussion on the improved integration of STM into VSOD.
This is followed by a discussion on the enhanced integration of STM into VSOD.
Subsequently, we introduce an effective fusion approach for the effective fusion of various feature types.
In addition, we adapt the commonly used multitask learning framework with edge supervision, a technique prevalent in ISOD, for application in VSOD.
Finally, we provide a detailed explanation of the training loss for the entire model.
\begin{figure*}[ht]
    \centering
    \includegraphics[width=\textwidth]{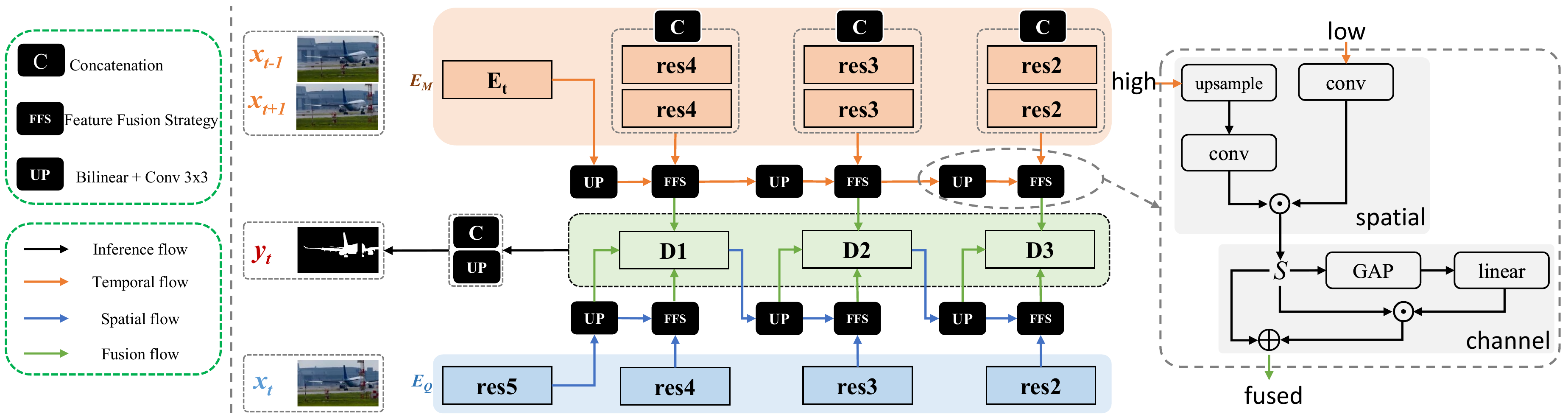}
    \caption{Detailed structure of the decoder. The high-level temporal feature is fused with the low-level features of adjacent frames by FFS, and the spatial branch follows the same process. The final saliency map is obtained by combining the spatiotemporal features generated in three stages. The channels of all features are shrunk to $64$ by $1\times1$ convolutional layer prior to fusion to reduce the computation costs.}
    \label{fig:decoder}
\end{figure*}
\subsection{Overall Architecture}\label{subsec:the-overall-architecture}
As shown in \cref{fig:model}, our model follows a U-shape~\cite{ronneberger2015u} encoder--decoder architecture.
In the encoding stage, given consecutive frame sequence $\textbf{X}=(x_1, x_2, \dots, x_T)$, when we process the current frame $x_t$, we utilize two parallel encodings, one to remember the information of the previous frame $x_{t-1}$ and next $x_{t+1}$, denoted as $E_M$, and the other to obtain the spatial information of the current frame, denoted as $E_Q$.
It should be noted that while $E_Q$ and $E_M$ share weights, they are labeled separately to facilitate a clearer description.
We employ ResNet~\cite{he2016deep}, which is pretrained on ImageNet~\cite{deng2009imagenet}, as our encoder.
ResNet generates five layers of features, which are denoted from low- to high-level as $conv1$, $res2$, $res3$, $res4$, and $res5$. $res5$ contains the most important salient object information, whereas other low-level features are used to reconstruct some detailed information, such as boundaries.
Our approach does not rely on external temporal information like optical flow.
Thus, the high-level features of the current frame $E_Q^{res5}$ and features $E_M^{res5}$ extracted from adjacent frames are sent to ASTM that generates the temporal salient high-level features $E_{t}$.

During the decoding process, the high-level temporal features $E_{t}$ and high-level spatial features $E_Q^{res5}$ are fused with their corresponding low-level features from $E_M$ and $E_Q$, respectively.
For instance, in the first stage, there are two branches: the temporal branch, fuses $E_{t}$ with $E_M^{res4}$ from adjacent frames, and the spatial branch, which combines $E_Q^{res5}$ with $E_Q^{res4}$.
Low- and high-level features in the two branches are fused by our proposed FFS\@.
Finally, the outputs of the two branches are merged through a $conv 3\times3$ layer to obtain the spatiotemporal saliency feature $D1$. $D2$ and $D3$ are also obtained through the above process.
We predict the final results from these stages.
The detailed structure of the decoder is shown in \cref{fig:decoder}.

\subsection{Adjacent Space-Time Memory Module}\label{subsec:adjacent-space-time-memory-module}
In VOS, STM is often inserted after $res4$ to generate temporal features used to reconstruct the predictions~\cite{oh2019video,seong2020kernelized,xie2021efficient,hu2021learning}.
They often discard $res5$ to reduce computational costs and maintain a large feature map size.
However, high-level features play a crucial role in finding salient objects, even to avoid smaller feature size, the use of ASPP~\cite{chen2017rethinking} instead of discarding high-level features is common in VSOD~\cite{zhang2021dynamic, fan2019shifting,yan2019semi}.
In the decoding stage, we consider how accurately the features can be upsampled with larger-sized low-level features, and we retain $res5$ to obtain salient cues.
In addition, given that the annotation of the first frame is provided in the semi-supervised VOS, the whole prediction process is naturally processed in chronological order.
When processing the current frame, they consider the previous frames and predictions, and the decoder follows a single-branch structure based on the output of STM when predicting the results.
By contrast, ASTM is only responsible for processing temporal information about the current frame from adjacent frames.

For VSOD, we extract temporal information from two adjacent frames that capture the most relevant object motion.
We avoid relying on memorizing previous predictions, which allows our model to operate without the need for two separate encoders for distinct inputs.
In the decoding stage, our approach is adapted to the commonly used two-branch structure in VSOD.

Specifically, we treat the current frame $x_{t}$ as \textbf{query} and the two adjacent frames $x_{t-1}$ and $x_{t+1}$ as \textbf{memory}.
These act as parallel inputs but share a common encoder.
Upon acquiring their high-level features $res5$, we utilize four distinct embedding layers for the query and memory frames to derive the corresponding \textbf{key} and \textbf{value}.
Key is used to match high-level semantic information, and value stores object saliency information.
${K}_{M}\in\mathbb{R}^{N\times H\times W \times C}$ and ${V}_{M}\in\mathbb{R}^{N\times H\times W \times 2C}$ denote the key and value from memory frames, respectively, where $H$ is the height, $W$ is the width, $C$ represents the channels, and $N$ is the number of memory frames.
Usually, {value} has twice the number of channels as the {key}.
Similarly, query frame outputs ${K}_{Q}\in \mathbb{R}^{ H\times W \times C}$ and ${V}_{Q}\in\mathbb{R}^{ H\times W \times 2C}$ for key and value, respectively.
The process can be formulated as follows:
\begin{equation}
\begin{aligned}
{K}_{Q},{V}_{Q} &= \mathit{Embedding}_{Q}\left(E_Q^{res5}\right)\\
{K}_{M},{V}_{M} &= \mathit{Embedding}_{M}\left(E_M^{res5}\right),
\end{aligned}\label{eq:embedding}
\end{equation}
where each $\mathit{Embedding}$ contains two different convolutional layers. $E_M^{res5}$ contains two frames, and the embedding results are concatenated along the $N$ dimension.

In the memory read process, we operate on relatively high-level features.
The smaller size of these features does not lead to computational challenges, allowing us to employ the memory read operation as described in~\cite{oh2019video}.
Initially, we calculate the similarity weights between ${K}_{Q}$ and ${K}_{M}$ in a non-local manner.
The $softmax$ function is then applied to normalize these weights.
Subsequently, the value is retrieved from memory and concatenated with ${V}_{M}$, forming our temporal salient high-level features$E_{t}$.
The entire process is articulated as follows:
\begin{equation}
\begin{aligned}
w &={K}_{M}\otimes{K}_{Q}  \\
\widetilde{w} &=\mathit{Softmax}\left(w\right)\\
E_{t} &= \left[\widetilde{w} \otimes {V}_{M},{V}_{Q}\right],
\end{aligned}\label{eq:read}
\end{equation}
where $\otimes$ denotes matrix multiplication, and $[\dots]$ denotes concatenation operation.

\begin{figure}[t]
	\centering
	\includegraphics[width=0.5\textwidth]{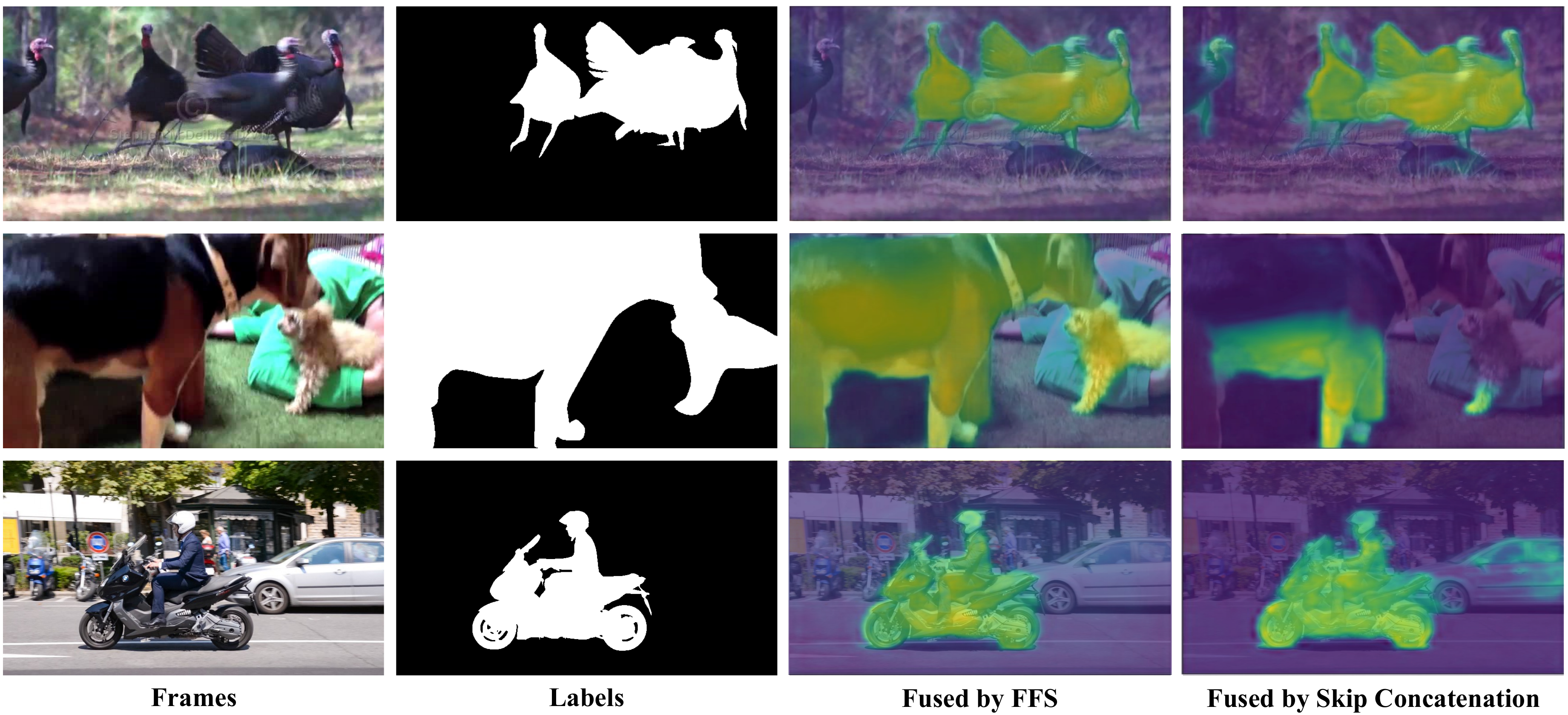}
	\caption{Visualization of three feature fusion cases. We compare the results of fusing high- and low-level features by the proposed FFS and skip concatenation. The features fused by FFS (column 3) have more accurate response and details for salient regions.}
	\label{fig:FFS}
\end{figure}
\begin{figure}[t]
	\centering
	\includegraphics[width=0.5\textwidth]{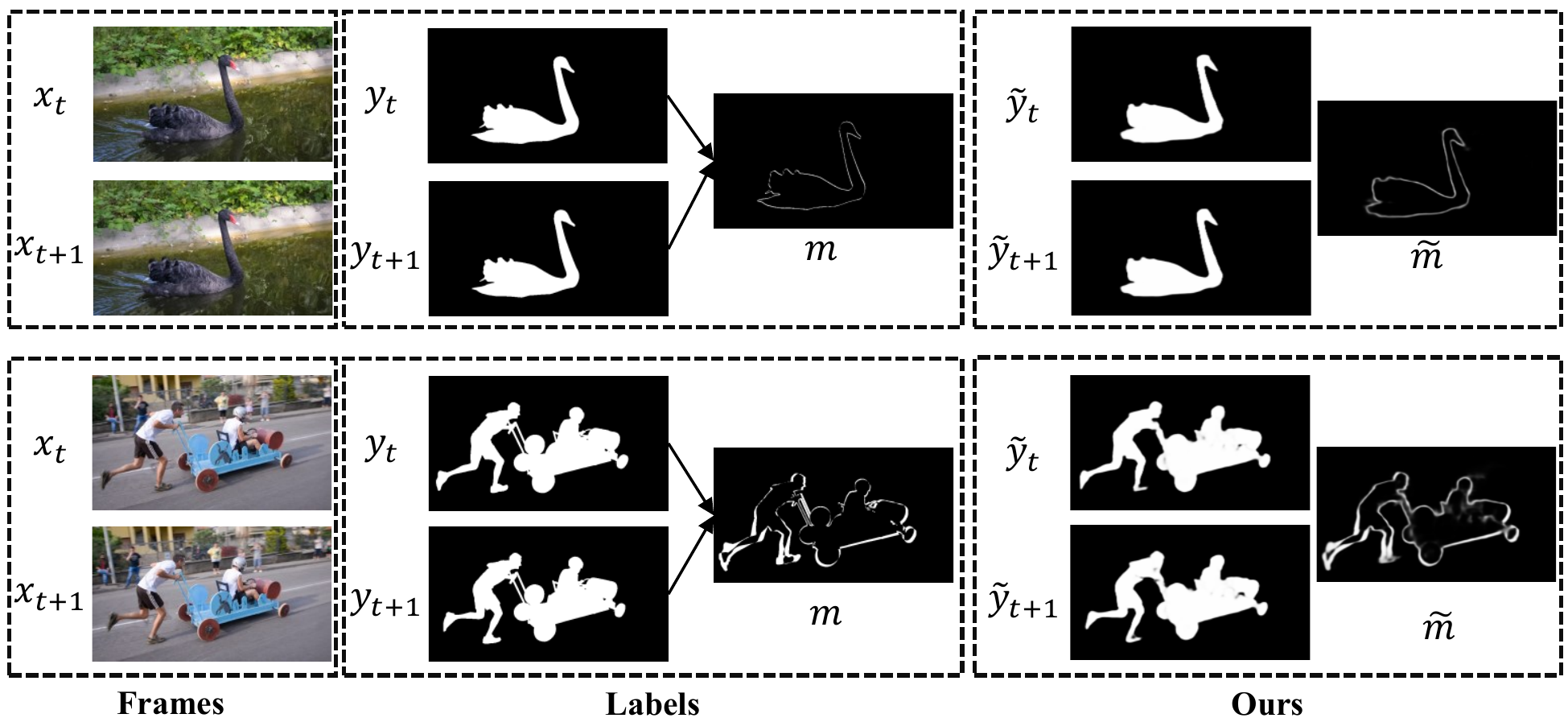}
	\caption{Process of generating motion-aware labels between frames. During model training, the two tasks share most features and promote each other. When the object is moving slightly (row 1), the boundary motion traces are also subtle, and the motion prediction part (leg of the man in row 2) is evident when the object moves quickly.}
	\label{fig:motion}
\end{figure}
\subsection{Feature Fusion Strategy}\label{subsec:feature-fusion-strategy}

The decoder consists of two branches.
The first branch fuses the spatial high-level features $E_Q^{res5}$ of the current frame with its corresponding lower-level features $E_Q^{res4}$, $E_Q^{res3}$ and $E_Q^{res2}$.
These lower-level features are crucial for reconstructing the details of the target object but may include background noise.
The second branch combines the temporal high-level features $E_t$ with the lower-level features of two adjacent frames.
This setup is illustrated in \cref{fig:decoder}.
We develop an effective yet straightforward feature fusion strategy that aims to accurately merge these features while minimizing the influence of irrelevant information, as high- and low-level features encapsulate different dimensions of information.

We represent the relatively higher and lower features as $h$ and $l$, respectively.
Initially, $h$ is upscaled to match the size of $l$ through bilinear interpolation followed by a convolutional layer.
Both $h$ and $l$ are then processed through separate convolution layers.
This step aims to reduce irrelevant noise in $l$.
The output channel of the former is set to $1$ to act as spatial attention weights, while the latter's output remains unaltered.
The multiplication fusion method~\cite{wei2020f3net,chen2020global} applied to these features effectively enhances the localization of target objects and suppresses extraneous information.
The fusion is achieved by element-wise multiplication of the two outputs, resulting in the spatial features $S$:
\begin{equation}
    \begin{aligned}
    \widetilde{h} &=\mathit{Upsampled}\left(h\right)  \\
    S &= \mathit{Conv_{c\rightarrow 1}}\left(\widetilde{h}\right)\times\mathit{Conv_{c\rightarrow c}}\left(l\right). \\
    \end{aligned}\label{eq:spatial}
\end{equation}
All features are uniformly compressed into $64$ channels using $1\times1$ convolutional layers before being fed into the decoder, a step designed to reduce computational costs.
Subsequently, these features undergo further stimulation in the channel dimension to amplify the channels with higher responses.
Each channel's weight interacts with its corresponding spatial feature, either enhancing or suppressing specific channel features to achieve improved overall results.
This process effectively balances the loss incurred from channel reduction without adding excessive parameters.
The final features $F$ can be obtained by
\begin{equation}
\begin{aligned}
c &=\mathit{Linear_{c\rightarrow c}}\left(\mathit{GAP}\left(S\right)\right)  \\
F &= S + c\times S. \\
\end{aligned}\label{eq:channel}
\end{equation}
Here, $GAP$ denotes global averaging pooling operation that squeezes the features to $\mathbb{R}^{ 1\times 1 \times C}$.
Subsequently, a linear layer is employed to produce the channel weights $c$.
The final results are obtained by weighting the spatial features $S$ with a residual connection.
\cref{fig:FFS} illustrates several examples of feature fusion cases.

In the initial stage of decoding, $E_Q^{res5}\left(h\right)$ and $E_Q^{res4}\left(l\right)$ are fused through the aforementioned process to derive the spatial feature $F_s^{1}$.
Concurrently, $E_t$ and $E_M^{res4}$ are processed to yield the temporal feature $F_t^{1}$.
It is important to note that $E_M^{res4}$ is formed by concatenating the low-level features from two adjacent temporal frames along the channel dimension.
The output $D_1$ of this first stage can be calculated as follows:
\begin{equation}\label{eq:eq5}
\begin{aligned}
D_{i} &= Conv_{3c\rightarrow c}\left[F_s^{i}, F_t^{i}, D_{i-1}\right],
\end{aligned}
\end{equation}
where $i$ denotes the decoding stage, $D_{0}$ is equivalent to $E_Q^{res5}$.
Following this, $D_{1}$ is fused with $E_Q^{res3}$, while in the temporal branch, a fusion occurs with $E_M^{res3}$ to produce $D_{2}$, as outlined in \cref{eq:eq5}.
The process is repeated to obtain $D_{3}$.
Ultimately, the saliency map $\widetilde{y}$ of the current frame is generated as follows:
\begin{equation}\label{eq:eq6}
\begin{aligned}
\mathbb{D} &= \left[D_{1},D_{2},D_{3}\right]\\
\widetilde{y} &= \operatorname{\sigma}\left(Conv_{3c\rightarrow 1}\left(\mathbb{D}\right)\right),
\end{aligned}
\end{equation}
where $\sigma$ indicates Sigmoid function that transforms the output into a probability map.

\subsection{Multitask learning with motion-aware loss}\label{subsec:multi-task-learning-with-motion-aware-loss}
Simultaneous learning coupled with edge detection is commonly practice in ISOD, as evidenced by various methods~\cite{PoolNet2019,wei2020label,ITSDNet2020,wang2019salient,li2018contour,wu2019stacked,su2019selectivity}.
In this context, ISOD and edge detection can share features derived from the encoder, facilitating multitask learning that mutually enhances each task's performance~\cite{DFInet2020}.
For example, edge detection aids in refining the boundaries of the target object within ISOD.
Under the assumption that objects exhibit only minor movements between adjacent frames, which essentially outline the contours of the object's motion, we propose a method to extract the moving parts of an object across frames.
Subsequently, we integrate multitask learning with VSOD after obtaining the contour information of the motion.
This approach represents a pioneering effort to incorporate edge supervision akin to that used in ISOD into VSOD, but from a temporal perspective.

Given two adjacent frames $x_{t}$, $x_{t+1}$, along with their corresponding labels $y_{t}$, $y_{t+1}$, the motion of the object is deduced using $m=\mathcal{XOR}\left(y_t, y_{t+1}\right)$.
Here, $\mathcal{XOR}$ signifies the exclusive or operation, which converts binary maps into new supervisory information, as illustrated in \cref{fig:motion}.
During the training phase for salient object prediction, the composite features $\mathbb{D}$, obtained from \cref{eq:eq5}, are temporarily stored.
These features are subsequently integrated with features from the succeeding frame to predict the motion $\widetilde{m}$.
The entire process can be described as follows:
\begin{equation}
\begin{aligned}
\widetilde{\mathbb{D}} &= \left[\mathbb{D}_t,\mathbb{D}_{t+1}\right]\\
\widetilde{m}  &= \operatorname{\sigma}\left(Conv_{6c\rightarrow 1}\left(\widetilde{\mathbb{D}}\right)\right).
\end{aligned}\label{eq:catD}
\end{equation}
Except for the final convolutional layers responsible for generating distinct predictions, these two tasks predominantly share the majority of their features.

To the best of our knowledge, this paper is the first to incorporate motion prediction into VSOD.
Previous methods have typically modeled inputs based on both appearance (the current frame) and motion (using optical flow or similar techniques).
However, their outputs have primarily focused on appearance predictions alone, manifested as the saliency map of the current frame.
This imbalance might inadvertently skew the model's focus, overly emphasizing appearance cues while potentially neglecting motion information.
In contrast, our method seamlessly integrates both appearance and motion predictions, as exemplified in \cref{fig:motion}.

\begin{table*}[h]

	\caption{Quantitative comparison of our model with other famous VSOD models. The top three metrics are highlighted in \textbf{bold}, \textcolor{red}{red}, and \textcolor{blue}{blue}. We use the three common metrics $\mathcal{S}$ (S-measure), $\mathcal{M}$ (MAE) and $\mathcal{F}$ (max F-measure) for comparison. $\uparrow$ indicates that larger is better, and $\downarrow$ indicates that smaller is better. All metrics are calculated on the test set of the following datasets. Unavailable metrics are denoted by /.}

	\renewcommand\arraystretch{1.3}
	\resizebox{\linewidth}{!}{
		\begin{tabular}{cc||cccccccccccccc||c}
			\toprule[1pt]
			\multicolumn{2}{c||}{\multirow{2}{*}{Metrics}}& SCNN & SCOM & DLVS & FGR & MBNM & PDB & MGA & RCR & SSAV  & PCSA  & WSV  & TENET & STVS  & DCF & Ours   \\
			&&\cite{tang2018weakly} &\cite{chen2018scom} &\cite{wang2017video} &\cite{li2018flow} &\cite{li2018unsupervised} &\cite{song2018pyramid} &\cite{li2019motion} &\cite{yan2019semi} &\cite{fan2019shifting}  &\cite{gu2020pyramid}  &\cite{zhao2021weakly}  &\cite{ren2020tenet} &\cite{chen2021exploring}  &\cite{zhang2021dynamic} &    \\  \hline
			\multicolumn{1}{c}{\multirow{3}{*}{\textbf{DAVIS}}}
			& $\mathcal{S}\uparrow$ & 0.783 & 0.832 & 0.802 & 0.838 & 0.887  & 0.882 & \textcolor{red}{0.910}   & 0.886 & 0.893 & 0.902 & 0.828 & \textcolor{blue}{0.905}  & 0.892 & \textbf{0.914}  & 0.897 \\
			& $\mathcal{M}\downarrow$ & 0.064 & 0.048 & 0.055 & 0.043 & 0.031  & 0.028 & 0.022   & 0.027 & 0.028 & 0.023 & 0.037 & \textcolor{blue}{0.021}  & 0.023 & \textbf{0.016}  & \textcolor{red}{0.020} \\
			& $\mathcal{F}\uparrow$ & 0.714 & 0.783 & 0.721 & 0.783 & 0.862  & 0.855 & \textcolor{blue}{0.892}   & 0.848 & 0.861 & 0.88  & 0.779 & \textcolor{red}{0.894}  & 0.865 & \textbf{0.900}  & 0.877 \\ \hline
			\multicolumn{1}{c}{\multirow{3}{*}{\textbf{FBMS}}}
			& $\mathcal{S}\uparrow$ & 0.794 & 0.794 & 0.794 & 0.809 & 0.857  & 0.851 & \textcolor{red}{0.908}   & 0.872 & 0.879 & 0.866 & 0.778 & \textbf{0.910}  & 0.872 & 0.873  & \textcolor{blue}{0.894} \\
			& $\mathcal{M}\downarrow$ & 0.095 & 0.079 & 0.091 & 0.088 & 0.047  & 0.064 & \textbf{0.027}   & 0.053 & 0.040 & 0.041 & 0.072 & \textbf{0.027}  & \textcolor{blue}{0.038} & 0.039  & \textcolor{red}{0.032} \\
			& $\mathcal{F}\uparrow$ & 0.762 & 0.797 & 0.759 & 0.767 & 0.816  & 0.821 & \textbf{0.903}   & 0.859 & 0.865 & 0.831 & 0.786 & \textcolor{red}{0.887}  & 0.854 & 0.840  & \textcolor{blue}{0.883} \\ \hline
			\multicolumn{1}{c}{\multirow{3}{*}{\textbf{SegV2}}}
			& $\mathcal{S}\uparrow$ & /     & 0.815 & 0.771 & 0.770 & 0.809  & 0.864 & 0.880   & 0.843 & 0.851 & 0.865 & 0.804 & /      & \textbf{0.894} & \textcolor{blue}{0.883}  & \textcolor{red}{0.886} \\
			& $\mathcal{M}\downarrow$ & /     & 0.030 & 0.048 & 0.035 & 0.026  & 0.024 & 0.027   & 0.035 & 0.023 & 0.025 & 0.033 & /      & \textcolor{blue}{0.016} & \textcolor{red}{0.015}  & \textbf{0.014} \\
			& $\mathcal{F}\uparrow$ & /     & 0.764 & 0.686 & 0.694 & 0.716  & 0.800 & 0.829   & 0.782 & 0.801 & 0.810 & 0.738 & /      & \textbf{0.864} & \textcolor{blue}{0.839}  & \textcolor{red}{0.850} \\ \hline
			\multicolumn{1}{c}{\multirow{3}{*}{\textbf{ViSal}}}
			& $\mathcal{S}\uparrow$ & 0.847 & 0.762 & 0.881 & 0.861 & 0.898  & 0.907 & 0.940   & 0.922 & 0.943 & 0.946 & 0.857 & 0.943  & \textbf{0.954} & \textcolor{red}{0.952}  & \textcolor{blue}{0.947} \\
			& $\mathcal{M}\downarrow$ & 0.071 & 0.122 & 0.048 & 0.045 & 0.020  & 0.032 & 0.017   & 0.027 & 0.020 & 0.017 & 0.041 & 0.021  & \textcolor{blue}{0.013} & \textbf{0.010}  & \textcolor{red}{0.012} \\
			& $\mathcal{F}\uparrow$ & 0.831 & 0.831 & 0.852 & 0.848 & 0.883  & 0.888 & 0.936   & 0.906 & 0.939 & 0.94  & 0.831 & \textcolor{blue}{0.947}  & 0.953 & \textbf{0.953}  & \textcolor{red}{0.948} \\ \hline
			\multicolumn{1}{c}{\multirow{3}{*}{\textbf{DAVSOD}}}
			& $\mathcal{S}\uparrow$ & 0.680 & 0.603 & 0.664 & 0.701 & 0.646  & 0.698 & 0.741   & 0.741 & 0.724 & 0.741 & 0.705 & \textcolor{red}{0.753}  & \textcolor{blue}{0.744} & 0.741  & \textbf{0.777} \\
			& $\mathcal{M}\downarrow$ & 0.127 & 0.219 & 0.129 & 0.095 & 0.109  & 0.116 & 0.083   & 0.087 & 0.092 & 0.086 & 0.103 & \textcolor{blue}{0.078}  & 0.086 & \textcolor{red}{0.074}  & \textbf{0.065} \\
			& $\mathcal{F}\uparrow$ & 0.541 & 0.473 & 0.541 & 0.589 & 0.506  & 0.572 & 0.643   & 0.653 & 0.603 & \textcolor{blue}{0.655} & 0.605 & 0.648  & 0.650 & \textcolor{red}{0.660}  & \textbf{0.708} \\ \hline
			\multicolumn{1}{c}{\multirow{3}{*}{\textbf{DAVSOD-N}}}
			& $\mathcal{S}\uparrow$ & 0.589 & /     & 0.599 & 0.638 & 0.597  & /     & /   & 0.674 & 0.661 & /     & 0.633 & /      & \textcolor{blue}{0.675}& \textcolor{red}{0.686}  & \textbf{0.688} \\
			& $\mathcal{M}\downarrow$ & 0.193 & /     & 0.147 & 0.126 & 0.127  & /     & /  & 0.118 & 0.117 & /     & 0.14  & /      & \textcolor{blue}{0.108} & \textcolor{red}{0.094}  & \textbf{0.088} \\
			& $\mathcal{F}\uparrow$ & 0.425 & /     & 0.416 & 0.468 & 0.436  & /     & /   & 0.533 & 0.509 & /     & 0.485 & /      & \textcolor{blue}{0.540}  & \textbf{0.574}  & \textcolor{red}{0.555} \\ \hline
			\multicolumn{1}{c}{\multirow{3}{*}{\textbf{DAVSOD-D}}}
			& $\mathcal{S}\uparrow$ & 0.533 & /     & 0.571 & 0.608 & 0.561  & /     & /   & \textbf{0.644} & 0.619 & /     & 0.572 & /      & \textcolor{red}{0.623} & 0.613  & \textcolor{blue}{0.622} \\
			& $\mathcal{M}\downarrow$ & 0.234 & /     & 0.128 & 0.131 & 0.14   & /     & /   & \textcolor{blue}{0.094} & 0.114 & /     & 0.163 & /      & 0.097 & \textcolor{red}{0.09}   & \textbf{0.089} \\
			& $\mathcal{F}\uparrow$ & 0.345 & /     & 0.336 & 0.39  & 0.352  & /     & /   & \textbf{0.444} & 0.399 & /     & 0.383 & /      & \textcolor{blue}{0.409} & 0.403  & \textcolor{red}{0.418} \\
			\bottomrule[1pt]
		\end{tabular}
	}

	\label{tab:1}
\end{table*}
\subsection{Loss Function}\label{subsec:loss-function}
Following previous works~\cite{qin2019basnet,ren2020tenet,zhang2021dynamic}, we use three loss functions to train our model, including binary cross entropy (BCE) loss, structural similarity (SSIM) loss~\cite{wang2004image} and intersection over union (IoU) loss~\cite{yu2016unitbox}.

BCE loss is the most important loss function in VSOD; it measures the distance between prediction and label in terms of binary classification of each pixel.
It can be formulated as follows:
\begin{equation}
\begin{split}
\begin{aligned}
\textit{L}_{bce} &=-\sum_{i=1}^{H} \sum_{j=1}^{W}\left[{Y}_{i j} \log \left(\widetilde{Y}_{ij}\right)+\left(1-{Y}_{i j}\right) \log \left(1-\widetilde{Y}_{ij}\right)\right],
\end{aligned}
\end{split}\label{eq:bce}
\end{equation}
where ${Y}$ is the label, and $\widetilde{Y}$ is the predicted saliency map.

Unlike BCE, which measures the loss based on independent pixels.
SSIM assesses the overall similarity between two images by using mean for luminance, variance for contrast, and covariance to measure structural similarity.
Specifically, several patches are initially generated using a sliding window of size $11\times11$.
The final SSIM loss is obtained by the following:
\begin{equation}
\textit{L}_{ssim}=1-\sum_{Y,\widetilde{Y}}^{N}\frac{\left(2 \mu_{Y} \mu_{\widetilde{Y}}+C_{1}\right)\left(2\sigma_{Y\widetilde{Y}}+C_{2}\right)}{\left(\mu_{Y}^{2}+\mu_{\widetilde{Y}}^{2}+C_{1}\right)\left(\sigma_{Y}^{2}+\sigma_{\widetilde{Y}}^{2}+C_{2}\right)},\label{eq:ssim}
\end{equation}
where $\mu$, $\sigma$, and $\sigma_{Y\widetilde{Y}}$ denote the mean, variance, and covariance, respectively.
By convention, $C_{1}=0.01^{2}$ and $C_{2}=0.03^{2}$ are selected to avoid dividing by 0.

IoU measures the overlap of the predicted salient objects with that of labels.
IoU loss is calculated by the following:
\begin{equation}
\textit{L}_{iou}=1-\frac{\sum_{i=1}^{H} \sum_{j=1}^{W} Y_{ij} \widetilde{Y}_{ij}}{\sum_{i=1}^{H} \sum_{j=1}^{W}[Y_{ij}+\widetilde{Y}_{ij}-Y_{ij} \widetilde{Y}_{ij}]} .\label{eq:iou}
\end{equation}

In addition, we predict the object motion, and the total loss function is defined as follows:
\begin{equation}
\label{eq:loss}
\textit{L}_{total}=\textit{L}_{bce}\left(\widetilde{y},y\right)+\textit{L}_{ssim}\left(\widetilde{y},y\right)+\textit{L}_{iou}\left(\widetilde{y},y\right)+\textit{L}_{bce}\left(\widetilde{m},m\right).
\end{equation}

%% file: 4experiments.tex
\section{Experiments}\label{sec:experiments}
We initially provide a brief introduction to the model training process, the datasets utilized in our experiments, and the evaluation metrics employed.
Subsequently, both quantitative and qualitative results are compared against popular VSOD models.
Additionally, ablation studies are conducted to underscore the effectiveness of our proposed method.

\subsection{Experimental Setup}\label{subsec:experimental-setup}
\subsubsection{Datasets}
We utilize several commonly-used datasets for our comparison experiments, including ViSal~\cite{wang2015consistent}, SegV2~\cite{li2013video}, FBMS~\cite{brox2010object}, DAVIS~\cite{perazzi2016benchmark}, and DAVSOD~\cite{fan2019shifting} to evaluate the effectiveness of our method.
\begin{itemize}
	\item \textbf{ViSal} is the first dataset released specifically for VSOD; it contains 17 videos with 193 frames.
	\item \textbf{SegV2} contains challenging scenarios including occlusion, shape distortion, and camera motion.
	This dataset is only used for testing.
	\item \textbf{FBMS} has 59 videos, but only 30 videos are used for testing because its labels are sparse and cannot satisfy our proposed learning from adjacent frames during training.
Some optical flow-based methods~\cite{wang2020cross,li2019motion} require only a single frame during training, typically utilizing 29 videos for training.
	\item \textbf{DAVIS} contains 30 videos for training and 20 videos for testing, which are available in 480~p and 1080~p resolutions; we use the former resolution.
	\item \textbf{DAVSOD} is currently the largest and most challenging VSOD dataset.
We use its training set and evaluate its default test dataset, which contains 35 videos. In addition, DAVSOD contains two harder test sets, \textbf{DAVSOD-Normal}~(DAVSOD-N) and \textbf{DAVSOD-Difficult}~(DAVSOD-D), which contain 25 and 20 videos, respectively. Compared to the training set, the two test sets contain extremely complex scenarios and multiple object saliency shift, which poses a significant challenge for future work.
\end{itemize}

We conduct evaluations on a total of seven different datasets.
Due to varying dataset distributions, none of the current state-of-the-art methods demonstrates optimal performance across all datasets.
This highlights that each method possesses its own unique strengths and weaknesses.
\begin{figure*}[t]
    \centering
    \includegraphics[width=\textwidth]{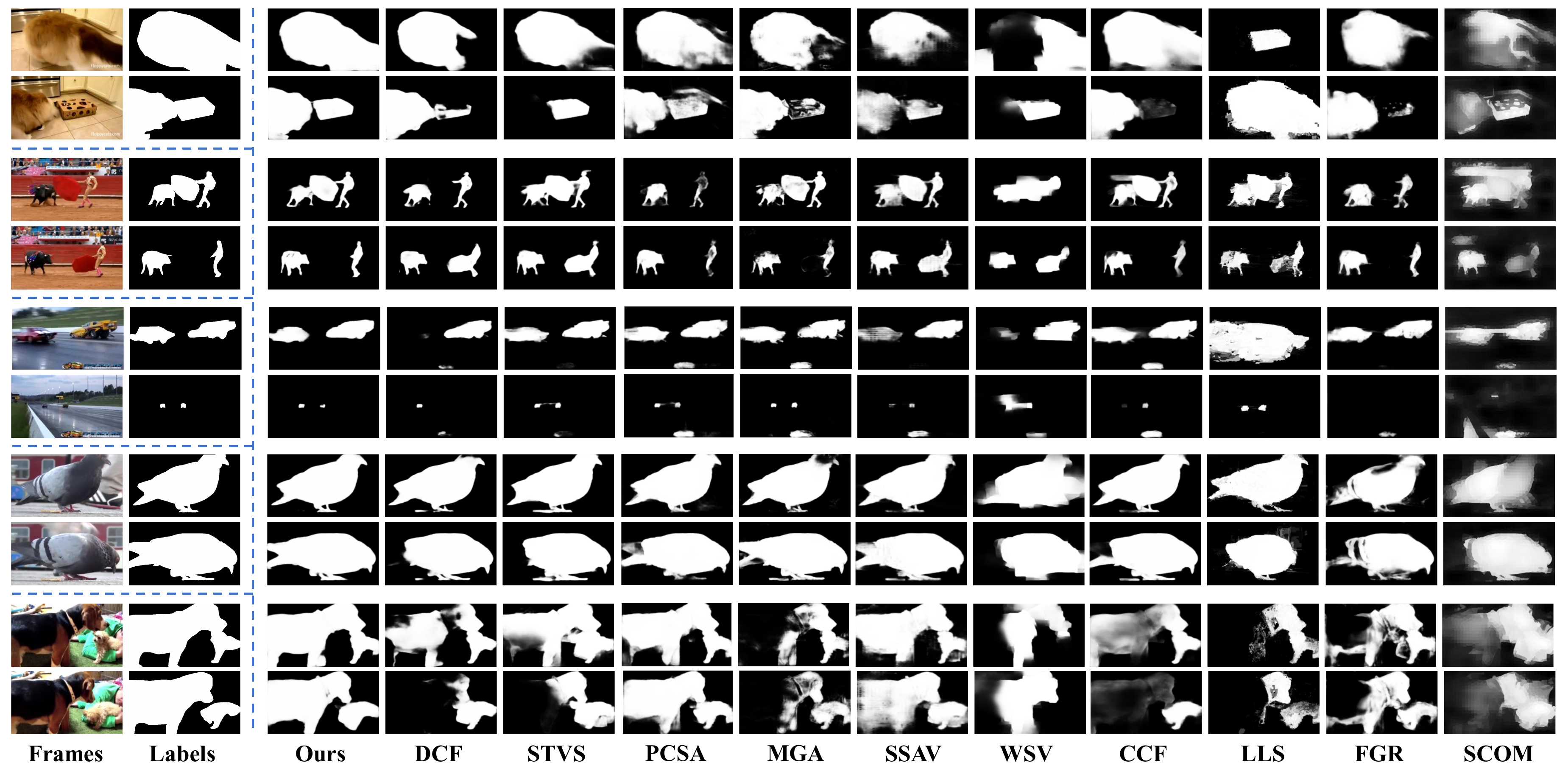}
    \caption{Qualitative comparison with the latest VSOD methods. We select some difficult scenarios, including saliency shift, multiple objects, fast moving object, and large object.}
    \label{fig:compare}
\end{figure*}
\subsubsection{Evaluation Metrics}
Consistent with previous methods, we used three commonly used evaluation metrics: mean absolute error ($\mathcal{M}$)~\cite{perazzi2012saliency}, structure measure S-measure ($\mathcal{S}$)~\cite{fan2017structure}, and F-measure (max $\mathcal{F}$)~\cite{achanta2009frequency}.

\subsubsection{Training Details}\label{subsubsec:traing}
Currently, the object categories in existing VSOD datasets are somewhat limited to effectively train a model with robust generalization capabilities.
For instance, the training sets of DAVIS and DAVSOD include only $91$ videos and feature a narrower range of object types compared to those in ISOD datasets.
In line with the common two-stage training approach, our model is initially pre-trained with ISOD task, followed by further training using the VSOD dataset.
Referring to the latest methods~\cite{chen2021exploring,zhang2021dynamic}, we utilize ResNet-101~\cite{he2016deep} pre-trained on ImageNet~\cite{deng2009imagenet} as the encoder.
All images are resized to $288\times288$.
The Adam optimizer is used to train the model.
The learning rate is set to $1e-5$ for the pre-trained parameters and $1e-4$ for the remaining parameters, with both learning rates decaying by half every $8$ epochs.
For data augmentation, images are randomly subjected to horizontal flipping and temporal inversion, each with a $50\%$ probability.

Initially, to augment the training process, we incorporate the DUTS dataset~\cite{wang2017learning} from the ISOD task, conducting the training for approximately 15 epochs until the model achieves convergence.
During this first stage, the motion-aware loss, specifically designed for videos, is omitted from the loss function.
Subsequently, we proceed to train the entire model on the training sets of DAVSOD and DAVIS, employing \cref{eq:loss} as the guiding loss function.
We set $T = 4$, a decision informed by previous methods\cite{zhang2021dynamic,zhao2021weakly,yan2019semi}.
This specific setting has been shown to effectively balance accuracy with computational costs.

\subsubsection{Testing Details}\label{subsubsec:testing}
During the testing phase, all generated saliency maps are upscaled to their original input dimensions using bilinear interpolation.
Our model achieves an inference speed of $0.04$ seconds per clip (four frames) on a single RTX 2080Ti GPU, achieving a rate of $100$ FPS.
Furthermore, in contrast to most previous models that are limited to a batch size of $1$ during testing, our model supports arbitrary batchsize.
It can process $18$ clips ($18\times 4=72$ frames) on a single RTX 2080Ti.
This batch processing takes approximately $0.44$ seconds per batch, resulting in an outstanding performance of 164 FPS.

\subsection{Comparisons with other methods}\label{subsec:comparisons-with-other-methods}
We conduct a comparative analysis of our method with 14 VSOD methods published in recent years, including DCF~\cite{zhang2021dynamic}, STVS~\cite{chen2021exploring}, TENET~\cite{ren2020tenet}, PCSA~\cite{gu2020pyramid}, SSAV~\cite{fan2019shifting}, RCR~\cite{yan2019semi}, MGA~\cite{li2019motion}, PDB~\cite{song2018pyramid}, MBNM~\cite{li2018unsupervised}, FGR~\cite{li2018flow}, DLVS~\cite{wang2017video}, SCOM~\cite{chen2018scom}, and SCNN~\cite{tang2018weakly}.
For most of these methods, results are obtained using their publicly available codes or models.
In cases where results are difficult to obtain, we directly cite the results from the respective papers.
Evaluation metrics are based on the commonly used tools referenced in~\cite{fan2019shifting}.

\subsubsection{Quantitative comparison}
We present three key metrics in \cref{tab:1} for all evaluated methods across the seven datasets.
It's important to note that each dataset has a unique distribution, and some are not specifically tailored for VSOD.
As a result, even the most recent models are unable to achieve the best metrics consistently across all datasets.
However, our model consistently ranks among the top three in most metrics.
This is particularly notable on the largest dataset, DAVSOD, where our model demonstrates a significant improvement over previous methods.

Among the latest methods, DCF and STVS, which are both non-optical flow models, exemplify the trend towards real-time VSOD.
The results indicate that STVS excels on the SegV2 dataset, while DCF leads on the ViSal and DAVIS datasets.
Our method also shows remarkable performance, achieving the best MAE on SegV2 and near-optimal results on ViSal and DAVIS.
Moreover, a noticeable limitation is observed with both DCF and STVS on the FBMS dataset.
This limitation is likely due to the absence of continuous labels in FBMS, which poses a challenge for non-optical flow methods in learning temporal information from consecutive frames.
Remarkably, our method, which does not use the FBMS training set, still achieves the second-best performance, trailing only behind optical flow-based methods that do utilize the FBMS training set.
This outcome highlights the strong generalization capability of our approach.

These datasets are either quite limited in size or not specifically tailored for VSOD.
DAVSOD, in contrast, encompasses a broader range of scenes and target objects, along with more intricate patterns of saliency shift and camera motion.
On its default test set, our method outperforms others by a significant margin.
We also obtained the best metrics on DAVSOD-N, with DCF being suboptimal.
Furthermore, as highlighted in \cref{tab:compareDCF}, the model size of DCF is nearly 40\% larger than ours.
The semi-supervised method RCR unexpectedly achieves the best metrics on the most difficult dataset, DAVSOD-D.
This is likely due to the poor quality of saliency maps generated by current methods on this dataset, introducing a degree of randomness in the results.
While our method achieves the lowest MAE on DAVSOD-D, it's important to note that the quality of our saliency maps on this dataset is significantly lower compared to other datasets.
Consequently, DAVSOD-D remains a formidable challenge that warrants further exploration in future research.

\begin{table}[t]
    \centering
		\caption{Runtime comparison with the latest methods. The data are derived from the original paper or its published code. We describe the input size, implementation framework, model size, running GPU, and FPS for all models.}
    \label{tab:compareDCF}
	\resizebox{\linewidth}{!}{
    \begin{tabular}{lcccccccc}
		\toprule[1pt]
		Methods		&Optical flow		&T 		&Input Size     		&Framework 		&Model(MB)		&GPU 		&FPS\\
		\midrule[0.5pt]
		Ours        &$\times$            &4 	&$288^2$        &Pytorch    	&194        	&RTX 2080TI		&100\\
		DCF~\cite{zhang2021dynamic}         &$\times$     			&4 		&$448^2$        &Pytorch    	&274        	&RTX 2080Ti		&28\\
		STVS~\cite{chen2021exploring}         &$\times$     			&3 		&$256^2$        &Pytorch    	&184        	&GTX 1080Ti		&50\\
		WSV~\cite{zhao2021weakly}         &$\checkmark$    		&4 		&$256^2$        &Pytorch    	&131        	&RTX 2080Ti		&29\\
		MGA~\cite{li2019motion}         &$\checkmark$    		&1 		&$512^2$        &Pytorch    	&350        	&GTX 1080Ti		&14\\
		RCR~\cite{yan2019semi}         &$\checkmark$    		&4 		&$448^2$        &Pytorch    	&206        	&GTX 1080		&27\\
		SSAV~\cite{fan2019shifting}         &$\times$    		&3 		&$473^2$        &Caffe    	&236        	&GTX TITANX		&20\\
		\bottomrule[1pt]

    \end{tabular}
	}
\end{table}

\begin{table}[t]
	\centering
	\caption{Comparison of methods that process four frames simultaneously, in terms of inference speed and MACs, evaluated with identical input dimensions and hardware specifications.}
	\label{tab:speed}
	\resizebox{\linewidth}{!}{
		\begin{tabular}{lcccccc}
			\toprule[1pt]
			Methods             		    & T 	& Input Size &Hardware	  & Speed(s)       & MACs(G) \\
			\midrule[0.5pt]
			Ours                            & 4     & $288^2$    &	i9-10900 \& RTX 2080Ti  & 0.04 			& 83 \\
			DCF~\cite{zhang2021dynamic}     & 4     & $288^2$    &	i9-10900 \& RTX 2080Ti	  & 0.09 			& 154  \\
			RCR~\cite{yan2019semi}          & 4     & $288^2$    &	i9-10900 \& RTX 2080Ti	  & 0.06   		& 182  \\
			WSV~\cite{zhao2021weakly}       & 4     & $288^2$    &	i9-10900 \& RTX 2080Ti	  & 0.11 			& 191  \\

			\bottomrule[1pt]

		\end{tabular}
	}
\end{table}

\subsubsection{Qualitative comparison}
To better demonstrate the effectiveness of our method, we conduct a visual comparison using saliency maps.
As depicted in \cref{fig:compare}, we selected five videos.
The first two videos feature scenarios with salient object shift, while the remaining three focus on the motion of identical salient objects.
Through this comparison, the superiority of our method becomes distinctly apparent.

The first video~(rows 1--2) captures a dog circling a box.
Initially, the box is obscured but later becomes visible; our method accurately detects both masks in these scenarios.
The second video~(rows 3--4) presents a more complex scenario of saliency shift.
It depicts a bullfighter provoking a bull with a red cloth, during which all elements become salient objects.
Subsequently, as the bullfighter removes the red cloth, the audience's focus noticeably shifts to the bullfighter and the bull.
Our method successfully identifies this salient object shift.
Among the other methods, only MGA captures this shift, but the quality of its results is notably inferior.

The third video~(rows 5--6) features two rapidly moving cars.
As the cars become smaller in the frame, other methods tend to mistakenly identify irrelevant small objects as salient.
In contrast, our method effectively identifies small objects without being misled by noise.
The fourth video~(rows 7--8) captures the dynamic body movements of a pigeon during feeding.
While other methods can accurately detect the pigeon only until it spreads its tail, our method adapts to these postural changes.
This is because our model concurrently predicts the moving parts of the body while detecting the salient object.
The last video~(rows 9--10) involves a black dog, a challenging subject for many models due to the dog's dark pixel values.
As the dog moves, the two most recent methods struggle to fully detect the black dog's body.
Our method, however, consistently excels in accurately identifying all objects, demonstrating its superior performance.

\subsubsection{Runtime Comparison}
\cref{tab:compareDCF} presents runtime information for several of the latest methods, with data extracted directly from the respective papers.
Our model significantly surpasses DCF~\cite{zhang2021dynamic} in performance, operating on the same GPU.
It's important to note that the input size of a model plays a critical role in determining its inference speed.
Even when accounting for the same input size, our model's inference speed is more than double that of DCF.
Additionally, despite WSV~\cite{zhao2021weakly} being a semi-supervised model with a smaller size and input dimensions compared to ours, its inference speed lags considerably behind.
This slower performance is attributed to the substantial computational demands imposed by its use of convLSTM.

To ensure a fair comparison, we selected methods that process four frames simultaneously, with each frame resized to $288\times288$ dimensions.
We conducted this assessment under uniform hardware conditions (CPU: i9-10900X@3.70GHz, GPU: RTX 2080Ti), measuring the inference speeds as detailed in \cref{tab:speed}.
Additionally, we provide the device-independent MACs metric for each iteration to evaluate computational efficiency.
In both inference speed and computational efficiency, our model demonstrates superior performance compared to these methods.

We also examine the underlying causes for the speed discrepancies among various methods.
DCF dynamically adjusts the parameters of convolution kernels during each operation, substantially diminishing its efficiency.
WAS employs convLSTM to manage temporal data, and such RNN-based methods inherently operate in a sequential manner, leading to extended runtimes.
RCR utilizes convGRU exclusively for high-level features.
While this reduces the execution speed to a degree, the computational load remains substantial.
Regarding optical flow-based methods, even discounting the optical flow computation time, the need for dual independent encoders — one for the image and another for the optical flow — hampers their efficiency.
This is why current methods no longer use optical flow.
In contrast, our model is inherently designed to concurrently process inter-frame information.
By avoiding reliance on multiple independent encoders for diverse input types during the encoding stage, we achieve a significant uptick in performance speed.

\begin{figure}
	\centering
	\includegraphics[width=0.5\textwidth]{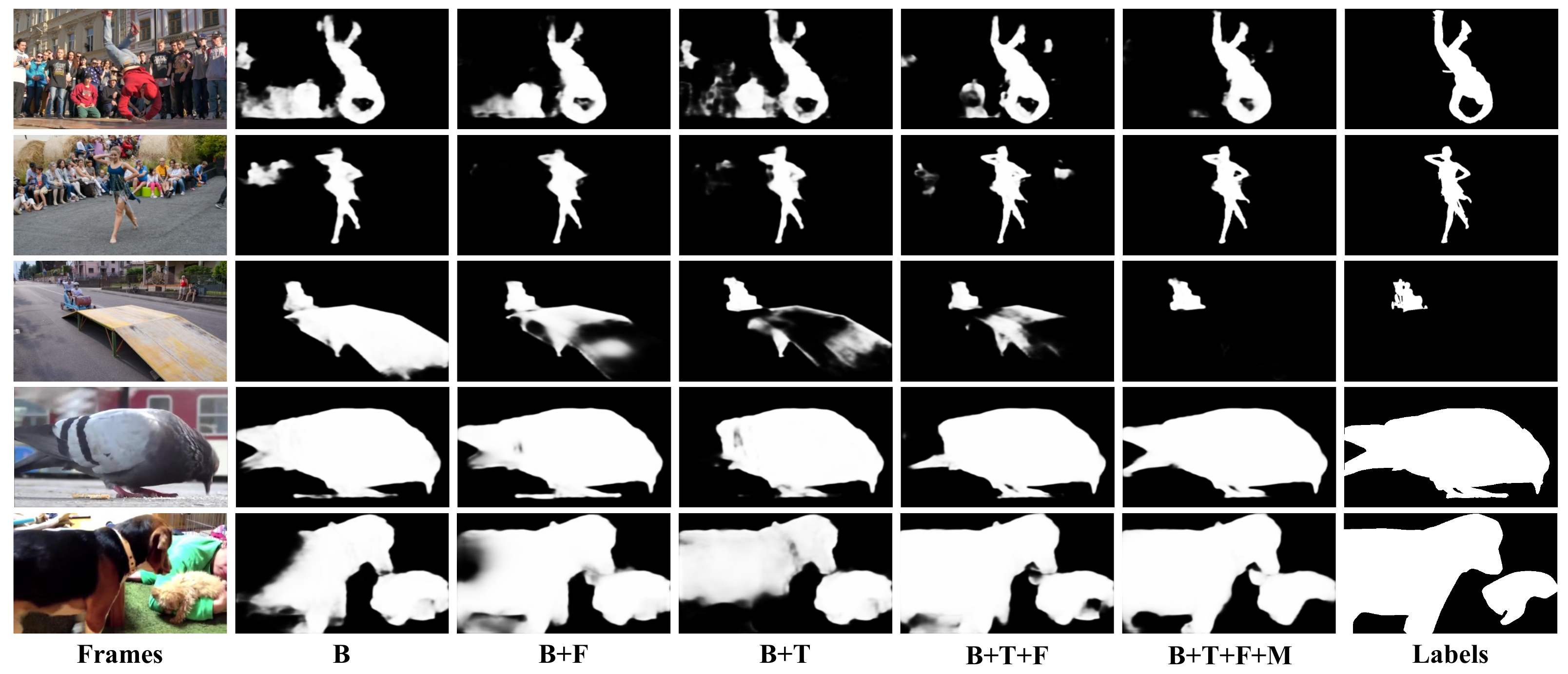}
	\caption{Qualitative comparison of the ablation study on our model. The visualization shows that the ability of our model becomes stronger with the help of the proposed components.}
	\label{fig:ablation}
\end{figure}
\begin{table}[t]
    \centering
    \caption{Ablation study on the four most commonly used VSOD datasets. `B' denotes the baseline model, and `B+F' denotes the baseline model with the proposed FFS, both of which are trained on ISOD dataset in stage 1. `B+T' denotes the model trained on VSOD after introducing ASTM in stage 2, and our complete model is denoted as `B+T+F+M'.}
    \label{tab:ablation}
\resizebox{\linewidth}{!}{
    \begin{tabular}{lcccccccc}
		\toprule[1pt]
		 \multicolumn{1}{c|}{\multirow{2}{*}{Methods}} & \multirow{2}{*}{stage 1} &\multirow{2}{*}{stage 2} & \multicolumn{3}{|c|}{DAVIS} & \multicolumn{3}{c}{FBMS} \\

		 \multicolumn{1}{c|}{}&  &   &\multicolumn{1}{|c}{$\mathcal{S}\uparrow$}  & $\mathcal{M}\downarrow$ & \multicolumn{1}{c|}{$\mathcal{F}\uparrow$} &$\mathcal{S}\uparrow$  & $\mathcal{M}\downarrow$ & $\mathcal{F}\uparrow$\\
		\midrule[0.5pt]
		B 		&		 \checkmark & 		&0.863		&0.033		&0.824		&0.846		&0.041		&0.790		\\
		B+F 	&		 \checkmark & 		&0.872		&0.029		&0.838		&0.852		&0.039		&0.802		\\
		B+T		&		  &\checkmark 		&0.889		&0.023		&0.862		&0.883		&0.033		&0.867		\\
		B+T+F   &		  &\checkmark 		&0.893		&0.022		&0.873		&0.896		&0.033		&0.881		\\
		B+T+F+M &		  &\checkmark 		&0.897		&0.020		&0.877		&0.894		&0.032		&0.883		\\
		\bottomrule[1pt]
    \end{tabular}
}
\newline
\vspace*{0.04 cm}
\newline
	\resizebox{\linewidth}{!}{
	    \begin{tabular}{lcccccccc}
		\toprule[1pt]
		 \multicolumn{1}{c|}{\multirow{2}{*}{Methods}} & \multirow{2}{*}{stage 1} &\multirow{2}{*}{stage 2} & \multicolumn{3}{|c|}{ViSal} & \multicolumn{3}{c}{DAVSOD} \\
		 \multicolumn{1}{c|}{}&  &   &\multicolumn{1}{|c}{$\mathcal{S}\uparrow$}  & $\mathcal{M}\downarrow$ & \multicolumn{1}{c|}{$\mathcal{F}\uparrow$} &$\mathcal{S}\uparrow$  & $\mathcal{M}\downarrow$ & $\mathcal{F}\uparrow$\\
		\midrule[0.5pt]
		B 		&		 \checkmark & 		&0.941		&0.014		&0.941		&0.736		&0.083		&0.647		\\
		B+F 	&		 \checkmark & 		&0.945		&0.013		&0.943		&0.749		&0.080		&0.661		\\
		B+T		&		  &\checkmark 		&0.943		&0.013		&0.936		&0.762		&0.073		&0.686		\\
		B+T+F   &		  &\checkmark 		&0.947		&0.012		&0.946		&0.768		&0.071		&0.693		\\
		B+T+F+M &		  &\checkmark 		&0.947		&0.012		&0.948		&0.777		&0.065		&0.708		\\
		\bottomrule[1pt]
    \end{tabular}
	}
\end{table}
\subsection{Ablation study}\label{subsec:ablation-study}
In this section, we perform comprehensive experiments to validate the effectiveness of our proposed modules.
The approach involves deconstructing the entire model into its basic architecture and incrementally integrating each proposed component, adhering to the outlined two-step training scheme.
Initially, we establish a baseline model using a vanilla encoder-decoder architecture, which employs feature concatenation (denoted as `B').
Prior to the integration of the ASTM module, our model lacks the capability to extract temporal information.
For clarity in the ablation study, we delineate the results according to the two stages of training: `stage 1' signifies training on the ISOD task, while `stage 2' refers to training on the VSOD task.

\subsubsection{Effectiveness of FFS}
FFS leverages semantic information from high-level features and enriches it with detailed object information from low-level features, thereby enhancing the model's ability to integrate features from different layers effectively.
The efficacy of FFS across both training stages is illustrated in \cref{tab:ablation}.

In stage 1, where the model is solely trained on the ISOD dataset, the baseline model shows limited performance on the four most commonly used VSOD datasets.
However, the inclusion of FFS significantly enhances the overall performance.
In stage 2, the model can be trained with VSOD datasets to learn temporal information by using ASTM.
During stage 2, the model, now trained with VSOD datasets, leverages the ASTM to learn temporal information.
It is observed that the model incorporating both ASTM and FFS (denoted as `B+T+F') outperforms the version trained only with ASTM (denoted as `B+T').
This is evident in \cref{fig:ablation}, where the integration of spatiotemporal multilevel features results in superior outcomes.
Therefore, the addition of FFS, whether in stage 1 or stage 2, aids the model in more accurately reconstructing final predictions and effectively mitigating background noise.
\begin{table}

    \centering
    \caption{Quantitative comparison of models with and without high-level features. $res_4$ indicates that the top-level features are discarded, and $res_5$ indicates that the top-level features are retained. Their results are equally poor in  stage 1, but the metrics of $res_5$ are improved significantly after stage 2.}
    \label{tab:res4}
\resizebox{\linewidth}{!}{
    \begin{tabular}{lcccccccc}
		\toprule[1pt]
		 \multicolumn{1}{c|}{\multirow{2}{*}{Methods}} & \multirow{2}{*}{stage 1} &\multirow{2}{*}{stage 2} & \multicolumn{3}{|c|}{DAVIS} & \multicolumn{3}{c}{FBMS} \\

		 \multicolumn{1}{c|}{}&  &   &\multicolumn{1}{|c}{$\mathcal{S}\uparrow$}  & $\mathcal{M}\downarrow$ & \multicolumn{1}{c|}{$\mathcal{F}\uparrow$} &$\mathcal{S}\uparrow$  & $\mathcal{M}\downarrow$ & $\mathcal{F}\uparrow$\\
		\midrule[0.5pt]
		$res_4$	&		 \checkmark & 		&0.865		&0.030		&0.836		&0.854		&0.042		&0.807		\\
		$res_5$ &		 \checkmark & 		&0.872		&0.029		&0.838		&0.852		&0.039		&0.802		\\
		$res_4$ &		  &\checkmark 		&0.880		&0.024		&0.856		&0.861		&0.036		&0.826		\\
		$res_5$ &		  &\checkmark 		&0.897		&0.020		&0.877		&0.894		&0.032		&0.883		\\
		\bottomrule[1pt]
    \end{tabular}
}
\newline
\vspace*{0.04 cm}
\newline
	\resizebox{\linewidth}{!}{
	    \begin{tabular}{lcccccccc}
		\toprule[1pt]
		 \multicolumn{1}{c|}{\multirow{2}{*}{Methods}} & \multirow{2}{*}{stage 1} &\multirow{2}{*}{stage 2} & \multicolumn{3}{|c|}{ViSal} & \multicolumn{3}{c}{DAVSOD} \\

		 \multicolumn{1}{c|}{}&  &   &\multicolumn{1}{|c}{$\mathcal{S}\uparrow$}  & $\mathcal{M}\downarrow$ & \multicolumn{1}{c|}{$\mathcal{F}\uparrow$} &$\mathcal{S}\uparrow$  & $\mathcal{M}\downarrow$ & $\mathcal{F}\uparrow$\\
		\midrule[0.5pt]
		$res_4$   &		 \checkmark & 		&0.942		&0.013		&0.938		&0.753		&0.080		&0.662		\\
		$res_5$	  &		 \checkmark & 		&0.945		&0.013		&0.943		&0.749		&0.080		&0.661		\\
		$res_4$	  &		  &\checkmark 		&0.936		&0.013		&0.939		&0.758		&0.076		&0.674		\\
		$res_5$   &		  &\checkmark 		&0.947		&0.012		&0.948		&0.777		&0.065		&0.708		\\

		\bottomrule[1pt]
    \end{tabular}
	}
\end{table}

\begin{figure}[t]
    \centering
    \includegraphics[width=0.48\textwidth]{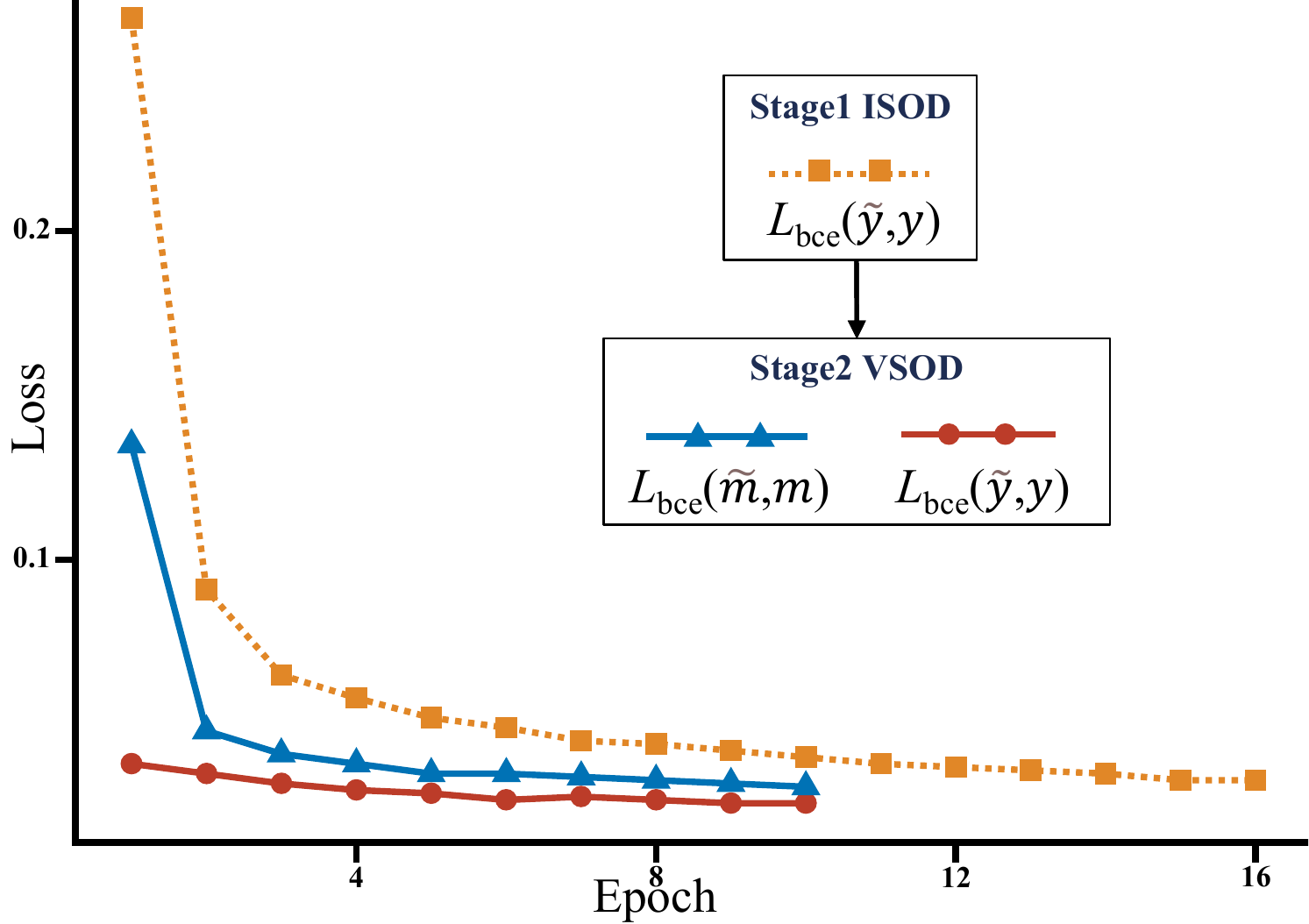}
    \caption{Training loss of our model throughout the two-stage training process. In stage 1, the model is pre-trained with ISOD task for about 15 epochs until convergence. In stage 2, the model is trained with the proposed motion-aware and its original loss on VSOD datasets, with both losses rapidly converging after 10 epochs.}
    \label{fig:loss}
\end{figure}
\subsubsection{Effectiveness of ASTM}
The limited quantity of existing VSOD datasets makes it challenging to directly train a VSOD model.
Consequently, almost all models initially train their spatial branch with the help of ISOD dataset, followed by the training of the temporal branch on the VSOD dataset.
In this experiment, we integrate ASTM as the temporal branch after the model pre-training on ISOD (denoted as `B+F'), and then proceed to train all branches on the VSOD dataset.
Notably, ASTM extracts information from adjacent frames, functioning in a manner akin to a bidirectional mechanism.
This contrasts with some optical flow-based methods that provide temporal information in a strictly chronological sequence.

The incorporation of ASTM significantly enhances the performance of our model (denoted as `B+T+F'), enabling it to surpass existing methods in certain metrics.
On both of the largest VSOD datasets, DAVIS and DAVSOD, the MAE metric is improved by $0.01$.
Remarkably, even without using the training set of FBMS, our model achieves an 18\% improvement in MAE.
on the relatively smaller ViSal dataset, our method yields satisfactory results by relying solely on the basic baseline architecture, without integrating temporal information.
This achievement highlights the exceptional efficacy of our network structure.

\subsubsection{Effectiveness of motion-aware loss}
As detailed in (\cref{subsec:multi-task-learning-with-motion-aware-loss}), motion prediction empowers the model to accurately predict the moving parts within salient regions.
As shown in \cref{fig:ablation}, the incorporation of motion prediction (denoted as `B+T+F+M') remarkably improves the model's ability to predict the complete objects.
Thanks to our multitask learning approach, which amalgamates VSOD and motion prediction, our method is capable of performing motion prediction during testing as well.
Even in certain high-speed scenes, such as vehicle driving, the motion prediction can predict accurately.
Additionally, the motion-aware loss exhibits rapid convergence during the VSOD training phase.
As illustrated in \cref{fig:loss}, in stage 2, these two tasks share the same feature layers, and both losses converge rapidly.

\begin{figure}
	\centering
	\includegraphics[width=0.5\textwidth]{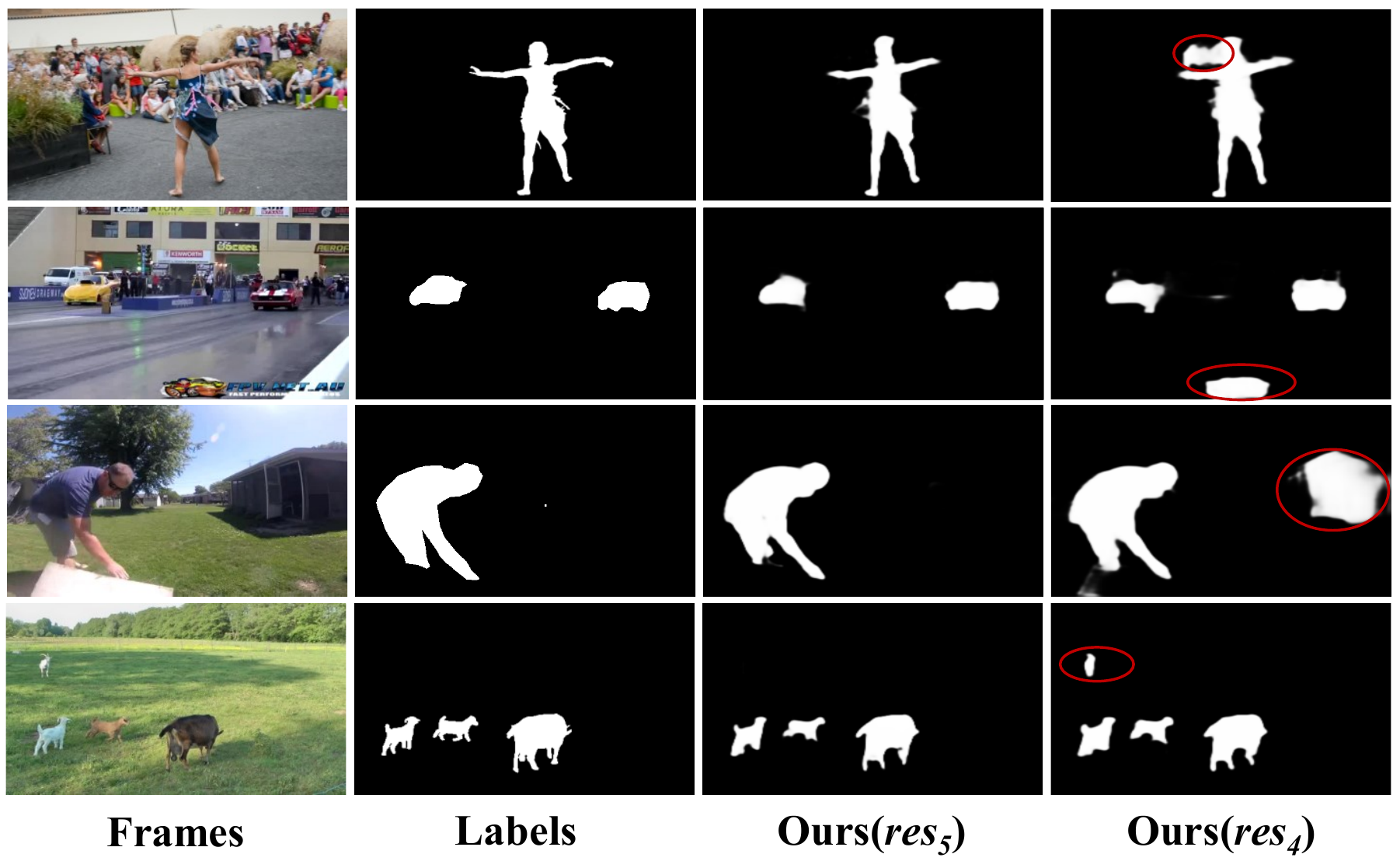}
	\caption{Comparison of results with and without high-level features. When the video has multiple objects, high-level features can better distinguish those that are salient or non-salient.}
	\label{fig:res4}
\end{figure}

\subsubsection{Importance of high-level features}
In the VOS task, the target object is predetermined, necessitating the segmentation of only this specified object in subsequent frames.
Consequently, to reduce computational costs and preserve a larger feature map size, high-level features are often omitted.
However, VSOD not only segments objects but also identifies salient ones, a process that benefits from the semantic information in high-level features.
We conducted a comparison between models using $res_5$ for high-level features and those using $res_4$, which is typical in VOS.
The results, as detailed in \cref{tab:res4}, show that in stage 1, when only trained on the ISOD dataset and evaluated on VSOD datasets, both models exhibit similar, yet modest performance metrics.
However, in stage 2, the model utilizing $res_5$ outperforms the one with $res_4$, particularly on larger dataset DAVSOD, where the salient objects undergo constant changes.

On the smaller ViSal dataset, where each video features a single, consistent object, the performance of the two models is comparable.
However, in scenarios involving multiple objects, as illustrated in \cref{fig:res4}, the model without the aid of $res_5$ encounters difficulties in differentiating between salient and non-salient objects.
This indicates the significant role of high-level features in complex object detection tasks.
\begin{table}

	\centering
	\caption{When the number of frames in the memory pool is increased, the computational efficiency of the model decreased by $20\%$, and there was no notable improvement in metrics. The model is prone to overfitting.}
	\label{tab:increaseT}
	\resizebox{\linewidth}{!}{
		\begin{tabular}{cccccccc}
			\toprule[1pt]
			\multicolumn{1}{c|}{\multirow{1}{*}{Memory}} & \multicolumn{1}{c}{\multirow{2}{*}{FPS}} & \multicolumn{3}{|c|}{DAVIS} & \multicolumn{3}{c}{FBMS} \\

			\multicolumn{1}{c|}{frames}&\multicolumn{1}{c}{}&  \multicolumn{1}{|c}{$\mathcal{S}\uparrow$}  & $\mathcal{M}\downarrow$ & \multicolumn{1}{c|}{$\mathcal{F}\uparrow$} &$\mathcal{S}\uparrow$  & $\mathcal{M}\downarrow$ & $\mathcal{F}\uparrow$\\
			\midrule[0.5pt]
			$x_{t-1}, x_{t+1}$	                &100          &0.897		&0.020		&0.877		&0.894		&0.032		&0.883		\\
			$x_{t-2}, x_{t-1},x_{t+1}, x_{t+2}$ &80($-20\%$)  &0.905		&0.019		&0.888		&0.873		&0.038		&0.856		\\
			\bottomrule[1pt]
		\end{tabular}
	}
	\newline
	\vspace*{0.04 cm}
	\newline
	\resizebox{\linewidth}{!}{
		\begin{tabular}{cccccccc}
			\toprule[1pt]
			\multicolumn{1}{c|}{\multirow{1}{*}{Memory}} & \multicolumn{1}{c}{\multirow{2}{*}{FPS}} & \multicolumn{3}{|c|}{ViSal} & \multicolumn{3}{c}{DAVSOD} \\

			\multicolumn{1}{c|}{frames}&\multicolumn{1}{c}{}&  \multicolumn{1}{|c}{$\mathcal{S}\uparrow$}  & $\mathcal{M}\downarrow$ & \multicolumn{1}{c|}{$\mathcal{F}\uparrow$} &$\mathcal{S}\uparrow$  & $\mathcal{M}\downarrow$ & $\mathcal{F}\uparrow$\\
			\midrule[0.5pt]
			$x_{t-1}, x_{t+1}$	                &100          &0.947		&0.012		&0.948		&0.777		&0.065		&0.708		\\
			$x_{t-2}, x_{t-1},x_{t+1}, x_{t+2}$ &80($-20\%$)  &0.943		&0.014		&0.945		&0.778		&0.064		&0.710		\\
			\bottomrule[1pt]
		\end{tabular}
	}
\end{table}
\begin{figure}[t]
	\centering
	\includegraphics[width=0.5\textwidth]{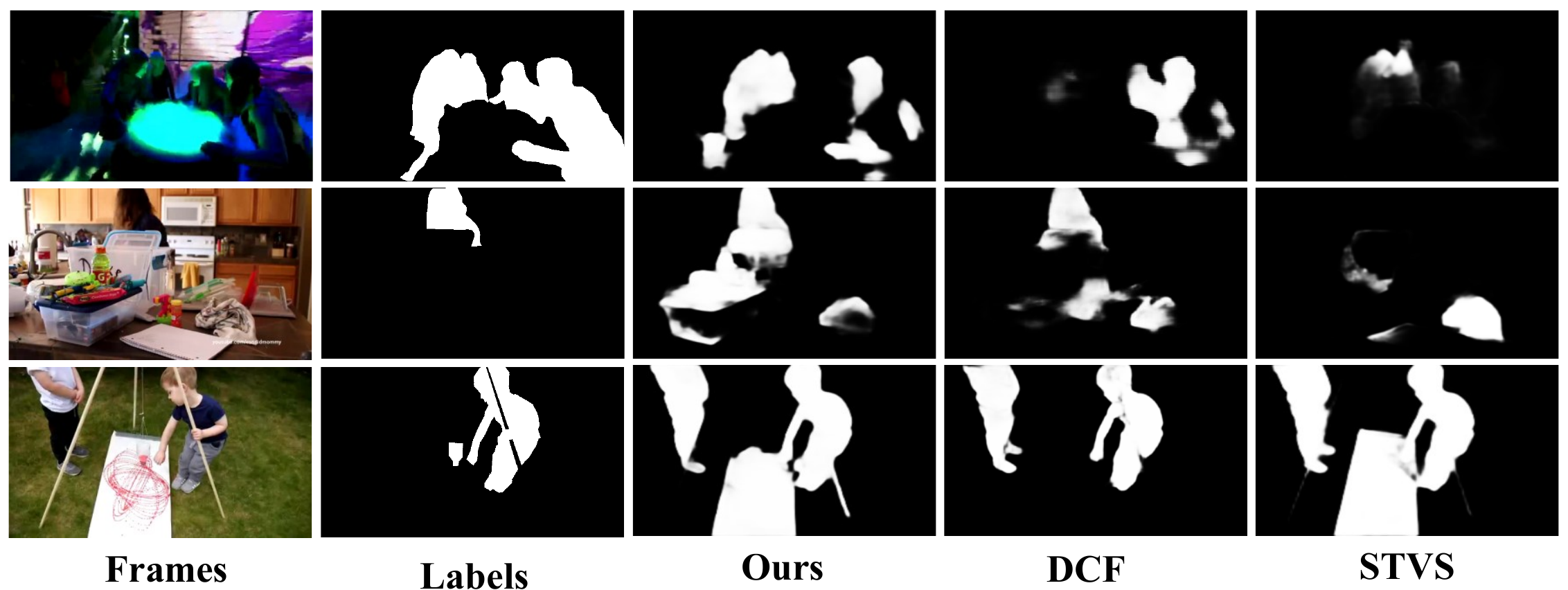}
	\caption{Some failure cases of our method, along with the results of the two most recent methods.}
	\label{fig:limitation}
\end{figure}
\subsubsection{Expanding the memory size}
We attempt to expand the number of temporal frames in the memory pool.
Our findings suggest that increasing the frame count does not proportionally enhance the model's performance, particularly when additional computational demands are taken into account.
\cref{tab:increaseT} details the results of these experiments, where we expand the sequence in the memory pool to include two frames before and two frames after the current frame.
The model's computational efficiency decreases by approximately $20\%$, but there is no corresponding improvement in performance metrics.

Previous method~\cite{gu2020pyramid} have indicated that increasing the temporal sequence does not significantly improve the segmentation effect of the VSOD model.
In our experiments, an extended temporal range increases the risk of model overfitting.
Considering these factors, solving the problem of long-term dependency is not as straightforward as simply expanding the temporal length.
Most current methods, including ours, operate within a 2 to 4 frame window.
This approach provides a better balance between computational efficiency and detection accuracy.
\subsection{Limitations}
We select some representative examples of failure cases, along with the results from the two most recent methods.
As illustrated in \cref{fig:limitation}, in scenes with complex lighting (row 1), the model struggles to distinguish the boundary between the object and the background.
In situations where the foreground contains a large, non-salient object and the target object is in the background (row 2), the model fails to discern depth information, resulting in an inability to accurately identify salient objects.
In scenarios involving long-term temporal dependencies, our model faces limitations due to memory or computational constraints, hindering its ability to process extremely long frames and consequently ascertain the saliency of the current object.
This often leads to failures in motion prediction as well.
Addressing these limitations represents a future direction for the development of VSOD\@.

%% file: 5conclusion.tex
\section{Conclusion}\label{sec:conclusion}

Distinct from previous methods, we integrate STM into VSOD to obtain temporal information, and subsequently adapt it to suit our specific requirements while maintaining high efficiency.
During the decoding stage, we design FFS to merge features at different levels using spatial and channel attention mechanisms.
This process effectively reconstructs the final saliency map by combing high-level salient features with detailed object attributes from lower-level features.
Additionally, drawing inspiration from the boundary supervision frequently employed in ISOD, we introduce a boundary motion prediction method, adapting it for multitask learning within VSOD\@.
Experiments demonstrate that this method yields impressive results, particularly on the largest dataset, DAVSOD.

Currently, the field of VSOD is confronted with several significant challenges.
A primary issue is the limited diversity of objects in existing datasets.
To address this, the urgent development of multiclass VSOD datasets is required, enabling models to be trained directly on these more diverse VSOD datasets.
Another significant concern is the tendency of existing models to focus on short-term temporal relationships and are unable to effectively capture long-term temporal dependencies.
In particular, models that utilize the Transformer architecture~\cite{vaswani2017attention} show promise in potentially overcoming the challenges associated with long-sequence dependencies.